\documentclass[10pt,twocolumn,letterpaper]{article}

\usepackage[pagenumbers]{cvpr} % To force page numbers, e.g. for an arXiv version

\usepackage{graphicx}
\usepackage{amsmath}
\usepackage{amssymb}
\usepackage{booktabs}
\usepackage{array}
\usepackage{multirow}
\usepackage{makecell}
\usepackage{bbding}
\PassOptionsToPackage{normalem}{ulem}
\usepackage{ulem}
\usepackage{arydshln}
\usepackage{xcolor}         % colors
\usepackage[latin9]{inputenc}
\usepackage[accsupp]{axessibility}

\usepackage[pagebackref,breaklinks,colorlinks]{hyperref}

\usepackage[capitalize]{cleveref}
\crefname{section}{Sec.}{Secs.}
\Crefname{section}{Section}{Sections}
\Crefname{table}{Table}{Tables}
\crefname{table}{Tab.}{Tabs.}

\def\cvprPaperID{3545} % *** Enter the CVPR Paper ID here
\def\confName{CVPR}
\def\confYear{2023}

\begin{document}

\title{Rethinking Optical Flow from Geometric Matching Consistent  Perspective}

\author{Qiaole Dong\footnotemark[1], Chenjie Cao\footnotemark[1], Yanwei Fu\footnotemark[2]\\
School of Data Science, Fudan University\\
{\tt\small \{qldong18,20110980001,yanweifu\}@fudan.edu.cn}
% \and
}
\maketitle

\renewcommand{\thefootnote}{\fnsymbol{footnote}}
\footnotetext[1]{ Equal contributions.}
\footnotetext[2]{ Corresponding author. The author is also with Shanghai Key Lab of Intelligent Information Processing, and Fudan ISTBI-ZJNU Algorithm Centre for Brain-inspired Intelligence, Zhejiang Normal University, Jinhua, China.}

%%%%%%%%% ABSTRACT
\begin{abstract}
Optical flow estimation is a challenging problem remaining unsolved. Recent deep learning based optical flow models have achieved considerable success. However, these models often train networks from the scratch on standard optical flow data, which restricts their ability to robustly and geometrically match image features. In this paper, we propose a rethinking to previous optical flow estimation. We particularly leverage Geometric Image Matching (GIM) as a pre-training task for the optical flow estimation (MatchFlow) with better feature representations, as GIM shares some common challenges as optical flow estimation, and with massive labeled real-world data.
% In this paper, we propose to use geometric image matching, which shares some common challenges with the optical flow estimation but with massive labeled real-world data, as a pre-training task for optical flow estimation (MatchFlow) to learn better representation for feature extraction.
Thus, matching static scenes helps to learn more fundamental feature correlations of objects and scenes with consistent displacements.
% rather than forming potentially confusing concept of 3D multi-object motion too early.
% Besides, matching static scenes helps to learn more fundamental factors of single object or scene's feature matching before prematurely forming potentially confounding priors for 3D multi-object motion. 
Specifically, the proposed MatchFlow model employs a QuadTree attention-based network pre-trained on MegaDepth to extract coarse features for further flow regression. Extensive experiments show that our model has great cross-dataset generalization. Our method achieves 11.5\% and 10.1\% error reduction from GMA on Sintel clean pass and KITTI test set. At the time of anonymous submission, our MatchFlow(G) enjoys state-of-the-art performance on Sintel clean and final pass compared to published approaches with comparable computation and memory footprint. Codes and models will be released in \url{https://github.com/DQiaole/MatchFlow}.
\end{abstract}

%%%%%%%%% BODY TEXT
\section{Introduction}
\label{sec:intro}

\begin{figure*}
\centering
\includegraphics[width=0.98\linewidth]{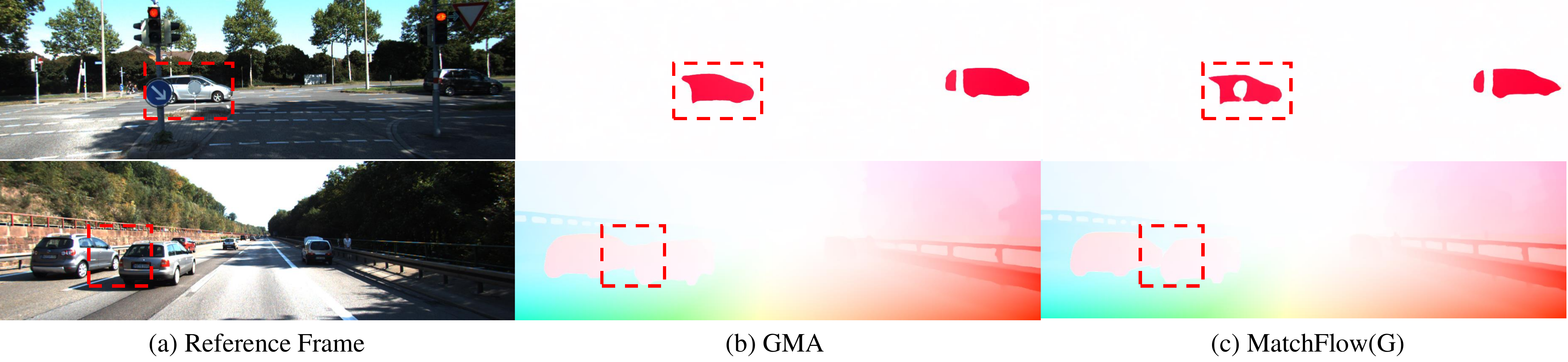}
\vspace{-0.1in}
\caption{Qualitative results on KITTI test set. Red dashed boxes mark the regions of substantial improvements. Please zoom in for details.\label{fig:kitti_qualitative}}
\vspace{-0.1in}
\end{figure*}

This paper studies optical flow estimation, which is the problem of estimating the per-pixel displacement vector between two frames. 
%which is a very hard problem in computer vision and remains challenging.
It is very useful to various real-world applications, such as video frame interpolation~\cite{jiang2018super}, video inpainting~\cite{gao2020flow}, and action recognition~\cite{sun2018optical}. 
The recent direct-regression based methods~\cite{dosovitskiy2015flownet, ilg2017flownet, sun2018pwc, teed2020raft, ranjan2017optical, jiang2021learning}  
have achieved great success by using powerful deep models, especially the recent transformers~\cite{xu2022gmflow, huang2022flowformer}. Among them, RAFT~\cite{teed2020raft} employs a convolutional GRU for iterative refinements, which queries local correlation features from a multi-scale 4D correlation volume. And GMA~\cite{jiang2021learning} further proposes a global motion aggregation module based on the self-similarity of image context, which greatly improves the performance within the occluded regions without degrading the performance in non-occluded regions. 
Typically, these
models often train networks from the scratch on standard
optical flow data, with the matching module (correlation volume) to help align the features of different images/frames. 
% However, 
% such a learning strategy can not empower their models with the ability of robustly and  geometrically matching image features. 
Generally, these current optical flow estimation algorithms still can not robustly handle several intractable cases, \textit{e.g.}, small and fast-moving objects, occlusions, and textureless regions, as these estimators have very restricted ability of robustly learning the local image feature correspondence of different frames.

% Since Horn and Schnuck~\cite{horn1981determining} first formulated optical flow estimation as a continuous optimization problem through a variational framework, optical flow estimation has long been regarded as an energy minimization problem, which optimizes an energy function composed of \emph{data} term and \emph{regularization} term. Although these methods~\cite{zach2007duality,chen2016full} can achieve some success, they have difficulty in robustly estimating optical flows across different motion distributions.

In this paper, we aim to provide a rethinking to the importance of Geometric
Image Matching (GIM) to the optical flow estimation.  In particular, 
despite GIM being designed to deal with the geometrically matching of static scenes, it indeed shares some common challenges with optical flow estimation, such as large displacement and appearance change~\cite{truong2020glu}. 
Thus, we advocate that the deep models for optical flow estimation should be trained from  
matching static scene pairs with consistent displacements. This can potentially help these models to learn the local low-level features and color correlations at the early stages of networks, before extracting the priors for 3D multi-object motion.
Furthermore, compared to optical flow data, it is much easier and simple to collect the real-world GIM data~\cite{dai2017scannet, li2018megadepth}, labeled by camera poses and depth computed from ground-truth or pioneering multi-view stereo manners~\cite{schonberger2016pixelwise}.  Such extensive real-world data largely improves the generalization of optical flow.

On the other hand, we can also emphasize a rethinking of the general training pipeline of optical flow. Specifically, since the creative work of FlowNet2~\cite{ilg2017flownet}, optical flow models~\cite{sui2022craft, jiang2021learning, sui2022craft} are often trained following the schedule of \textit{Curriculum Learning}~\cite{bengio2009curriculum}, \emph{i.e.}, from FlyingChair~\cite{dosovitskiy2015flownet} to FlyThings3D~\cite{mayer2016large} and finally to Sintel~\cite{butler2012naturalistic} or KITTI~\cite{geiger2012we}.
Nevertheless, the motion contained in FlyingChair is still far from the simplest scenario. Empirically, the static scene under viewpoint changes of GIM~\cite{sun2021loftr, tang2022quadtree}, can also be taken as one special type of optical flow estimation, which is even much simpler than FlyingChair. Therefore, it is reasonable to take GIM amenable for being the very first stage of the curriculum learning pipeline. Essentially, as mentioned above, the GIM data can be easily collected at large-scale, and thus will greatly benefit the learning of deep models.

% And we find that optical flow estimation of , which is also known as ~\cite{sun2021loftr, tang2022quadtree}, is much simpler than FlyingChair and more suited for being the first stage of Curriculum Learning. 

% large-scale datasets also play an important role in the deep learning based optical flow estimation. 
% Since the creative work of FlowNet2~\cite{ilg2017flownet}, optical flow models~\cite{sui2022craft, jiang2021learning, sui2022craft} are often trained following the schedule of Curriculum Learning~\cite{bengio2009curriculum}, \emph{i.e.} from FlyingChair~\cite{dosovitskiy2015flownet} to FlyThings3D~\cite{mayer2016large} and finally to Sintel~\cite{butler2012naturalistic} or KITTI~\cite{geiger2012we}.
% However, the motion contained in FlyingChair is still far from the simplest scenario. 

% And we find that optical flow estimation of static scene under viewpoint changes, which is also known as Geometric Image Matching (GIM)~\cite{sun2021loftr, tang2022quadtree}, is much simpler than FlyingChair and more suited for being the first stage of Curriculum Learning. Ideally, 

% Furthermore,  can help networks learn the fundamental features and color correlations,

% rather than forming potentially confusing priors for 3D multi-object motion too early. 

% Hence the image matching based pre-training can be well generalized to optical flow.
% Furthermore, a static scene can help networks learn the fundamentals of color/feature matching without prematurely forming potentially confusing priors for 3D multi-object motion. 

 % And we elaborate on this idea (MatchFlow) in this paper.

Formally, this paper well elaborates on the key idea of taking GIM as the prefixed task for optical flow. We draw the inspiration from recent GIM works, and present a novel MatchFlow that can effectively generalize the pre-trained image matching module to estimate optical flow. The key component of MatchFlow is a new module of Feature Matching Extractor, composed of Resnet-16 and 8 interleaving self/cross-Quadrtee attention blocks, trained on GIM task and used to get the 4D correlation volume. After Feature Matching Extractor, our MatchFlow still takes the common practice of using GRUs module to handle optical flow estimation, with an optional GMA modelling the context information. In this paper, we denote the full model based on GMA as \textbf{MatchFlow(G)}, and the model without GMA module (a.k.a. RAFT) as \textbf{MatchFlow(R)}.

%
%
% matching and optical flow, relationship. 
%
% optical flow training pipeline, curriculum learning --->
%  missing image matching 
%
% --> ideally, matching can be taken as a preluded/prefixed task for optical flow. We elaborate this idea in this paper.
%
% On the other hand, self-supervised pre-training representation learning~\cite{Radford2018ImprovingLU,Radford2019LanguageMA,Brown2020LanguageMA,Devlin2019BERTPO} has benefited Natural Language Processing (NLP) fields a lot. The success of pre-training on NLP also stimulates the development of Computer Vision. Lots of representation learning models~\cite{He2021MaskedAA,Wei2021MaskedFP,bao2021beit,tong2022videomae} based on the \emph{masking-and-predicting} strategy have achieved great success on downstream tasks of image and video.

% Therefore, in this paper, we propose using GIM as a pre-training task for optical flow estimation (MatchFlow). 
% Furthermore, few investigations about how to improve the \emph{feature representation learning} for optical flow estimation have been confirmed.

Following the standard optical flow training procedure~\cite{teed2020raft, jiang2021learning}, we conduct extensive experiments on FlyingChair~\cite{dosovitskiy2015flownet}, FlyingThings3D~\cite{mayer2016large},  Sintel~\cite{butler2012naturalistic}, and KITTI~\cite{geiger2012we}.
Experiments results show that MatchFlow enjoys good performance and great cross-dataset generalization.  Formally, RAFT-based MatchFlow(R) obtains an Fl-all error 13.6\% on KITTI training set after being trained on synthetic datasets. In addition, GMA-based MatchFlow(G) achieves 11.5\% and 10.1\% error reduction from GMA on the Sintel clean pass and KITTI test set. The qualitative comparison on KITTI also shows the superior performance of MatchFlow as in Fig.~\ref{fig:kitti_qualitative}. Ablation studies verify that GIM pre-training indeed helps to learn better feature representation for optical flow estimation. 

We highlight our contributions as follows. 
(1) We reformulate the optical flow pipeline, and propose the idea of employing GIM as the preluding task for optical flow. This offers a rethinking to the learning based optical flow estimation.
%
%We highlight the underlying connection between these two tasks. 
(2) We further present a novel matching-based optical flow estimation model -- MatchFlow, which has the new module of Feature Matching Extractor, learned by the GIM pre-training task for optical flow estimation.  Accordingly, the pipeline of curriculum learning has also been updated to effectively train our MatchFlow.
(3) We introduce the massive real-world matching data to train our model. And thus 
our model can extract robust features to handle with the consistent motion of scenes, and common challenges faced by both tasks.
%
% Benefited from massive real-world matching data, consistent motion of scenes, and common challenges faced by both tasks, the matching based pre-training can help feature extractors to learn more robust feature representation for optical flow estimation. 
(4) We conduct extensive experiments and ablation studies to show that both the matching based pre-training and interleaving self/cross-attention modules are critical for the final optical flow performance. The proposed model shows great cross-dataset generalization and better performance over several optical flow competitors on several standard benchmarks.
%  also known as camera motion, 
% our proposed pre-training shows great cross-
% dataset generalization and can improve the base model on Sintel clean pass and KITTI a lot.

\section{Related Work}

%\subsection{Optical Flow Estimation as Energy Minimization}
\noindent\textbf{Optical Flow Estimation.}
Traditionally, optical flow estimation~\cite{horn1981determining, black1993framework, black1996robust, zach2007duality, revaud2015epicflow, brox2004high, bruhn2005lucas} is treated as an energy minimization problem.
% which optimizes an energy function composed of \emph{data} term and \emph{regularization} term. The data term and regularization term corresponds to the brightness constancy and spatial smoothness hypothesis respectively. 
%
% Horn and Schnuck~\cite{horn1981determining} first formulated optical flow estimation as a continuous optimization problem by a variational framework. 
% In order to conquer the over-smoothing and noise sensitivity of optical flow estimation algorithm, Black and Anandan~\cite{black1993framework} proposed a robust estimation framework. Zach \emph{et al.}~\cite{zach2007duality} further use L1 loss as data term to better handle outliers.
% And coarse-to-fine strategy~\cite{brox2004high, bruhn2005lucas} is proposed to first estimate a coarse flow in low resolution to handle large displacements. However, such methods are still inferior to estimate small fast-moving objects accurately and inability to remedy large mistakes from coarse stage completely.
%
%coarse-to-fine strategy still suffers from some dilemmas: it can't
%
%
% \noindent \textbf{Learning based Optical Flow Estimation.}
Nowadays, deep models~\cite{dosovitskiy2015flownet, ilg2017flownet, sun2018pwc, hui2018liteflownet} formulate the optical flow estimation as a regression problem
with an end-to-end trainable Convolutional Neural Network.
% They propose a large synthetic optical flow dataset for alleviating the problem of scarce real-world optical flow data and achieve a promising performance. Afterward, FlowNet2~\cite{ilg2017flownet} further proposes stacking multiple FlowNet for residual refinement and using a better training schedule, which first exceeds traditional optical flow estimation methods.
% Some  learning based algorithm also introduces the traditional coarse-to-fine strategy. PWC-Net~\cite{sun2018pwc} and Lite-FlowNet~\cite{hui2018liteflownet} construct a feature pyramid and utilize feature warping and cost volume for residual flow refinement.
%
More recently, deep learning based iterative refinement for optical flow~\cite{sun2018pwc,hui2018liteflownet, teed2020raft, jiang2021learning, sui2022craft, zhao2022global} has resulted in a big breakthrough. %originating from the creative work -- FlowNet2~\cite{ilg2017flownet}. 
% Afterward, IRR~\cite{hur2019iterative} proposes sharing the network weights between different iterations.
RAFT~\cite{teed2020raft} constructs a 4D multi-scale correlation volume and utilizes a convolutional GRU block as an update operator. And GMA~\cite{jiang2021learning} further proposes a global motion aggregation module to tackle the occlusion problem.  Besides, recent methods~\cite{luo2022learning,luo2022kpa,zheng2022dip,sun2022skflow,huang2022flowformer} mainly focus on the recurrent decoder of RAFT. These methods are orthogonal to ours as we mainly focus on GIM pretraining of feature encoder and provide a rethinking for optical flow from the geometric matching consistent scene perspective.

\noindent \textbf{Geometric Image Matching}.
Geometric image matching~\cite{lowe2004distinctive, revaud2019r2d2, sarlin2020superglue} tends to find correspondence among images with different views. Different from optical flow estimation, image matching usually assumes that the scene is static and the geometric displacements are due to the change of viewpoints (camera pose). 
% Most traditional matching methods~\cite{lowe2004distinctive, revaud2019r2d2, sarlin2020superglue} are composed of the pipeline as feature detection, description, and matching. 
% Recently, detector-free matching methods~\cite{choy2016universal} discard the stage of feature detection~\cite{lowe2004distinctive} and learn the dense correspondences directly~\cite{rocco2018neighbourhood, sun2021loftr, tang2022quadtree, chen2022aspanformer}. 
Among recent detector-free matching methods~\cite{choy2016universal,rocco2018neighbourhood, sun2021loftr, tang2022quadtree, chen2022aspanformer}, 
% LoFTR~\cite{sun2021loftr} utilizes a Transformer-based architecture to achieve global consistency between matches. 
Tang~\etal~\cite{tang2022quadtree} propose QuadTree attention to capture both fine-level details and long-range dependencies, which outperforms the Linear attention~\cite{katharopoulos2020transformers} used in~\cite{sun2021loftr}. In this work, we also employ a QuadTree attention-based network as our feature extractor to learn feature correlation between two-view images. Benefited from the matching pre-training, we can get much better flow estimation within non-occluded regions. Furthermore, global motion aggregation module~\cite{jiang2021learning} can help propagate the accurate flow within non-occluded regions to more challenging occluded ones, which can boost the performance of optical flow estimation a lot.

Besides, there is also a line of works~\cite{truong2020glu, truong2020gocor, zhao2022global, xu2022gmflow, Aleotti2021Learning} which try to reduce the gap between optical flow estimation and GIM. GMFlow~\cite{xu2022gmflow} formulates the optical flow estimation as a dense global matching problem, while GMFlowNet~\cite{zhao2022global} introduces the matching loss and matching initialization into the optical flow. On the other hand, Depthstill~\cite{Aleotti2021Learning} constructs matching pairs based on estimated depth for direct optical flow training and achieves superior generalization to unseen real data. 
We should clarify that our work is different from them because MatchFlow focuses on the GIM based pre-training for better feature representations from the perspective of curriculum learning~\cite{bengio2009curriculum}. Moreover, MatchFlow outperforms these competitors with superior generalization on Sintel.
% GMFlowNet is trained on flow dataset with matching loss directly, while we propose to first learn simpler single object matching.

\noindent \textbf{ Curriculum Learning}.
Large datasets are one of the driving forces of deep learning. However, due to it's difficult to provide reliable ground truth optical flow labels, there is limited target data. Thus optical flow models typically rely on the pre-training on the synthetic FlyingChair~\cite{dosovitskiy2015flownet} and FlyingThings3D~\cite{mayer2016large} dataset with the curriculum learning~\cite{bengio2009curriculum}. 
In contrast, we give a rethinking to this pipeline and propose using GIM as a pre-training task for optical flow.

% pre-training on two massive synthetic datasets-- FlyingChair and FlyingThings3D to learn a good representation. Besides, FlowNet2~\cite{ilg2017flownet} firstly discovery that training optical flow models on a series of datasets with increasing difficulty (from FlyingChair to FlyingThings3D), \emph{i.e.} curriculum learning~\cite{bengio2009curriculum}, can lead to dramatic performance improvement. 

% Inspired by this, we 

% propose using GIM as a pre-training task for optical flow, which can help learn simpler static scene matching with consistent displacements on massive real-world matching data~\cite{dai2017scannet, li2018megadepth} first before learning harder multi-object motion. 
% And AutoFlow~\cite{sun2021autoflow} further learn to render data for optical flow that optimizes the performance of a model on a target dataset. Although these synthetic datasets can boost the performance, they ....

\section{Methodology}
% \subsection{Overview}

\noindent\textbf{Problem Setup.} Given a pair of consecutive images, $I_1, I_2$, optical flow estimation is the task of estimating per-pixel dense displacement field $(f^1, f^2)$ which maps the location $(u, v)$ in $I_1$ to $I_2$ as $(u+f^1(u), v+f^2(v))$.

\begin{figure*}
\centering
\includegraphics[width=0.95\linewidth]{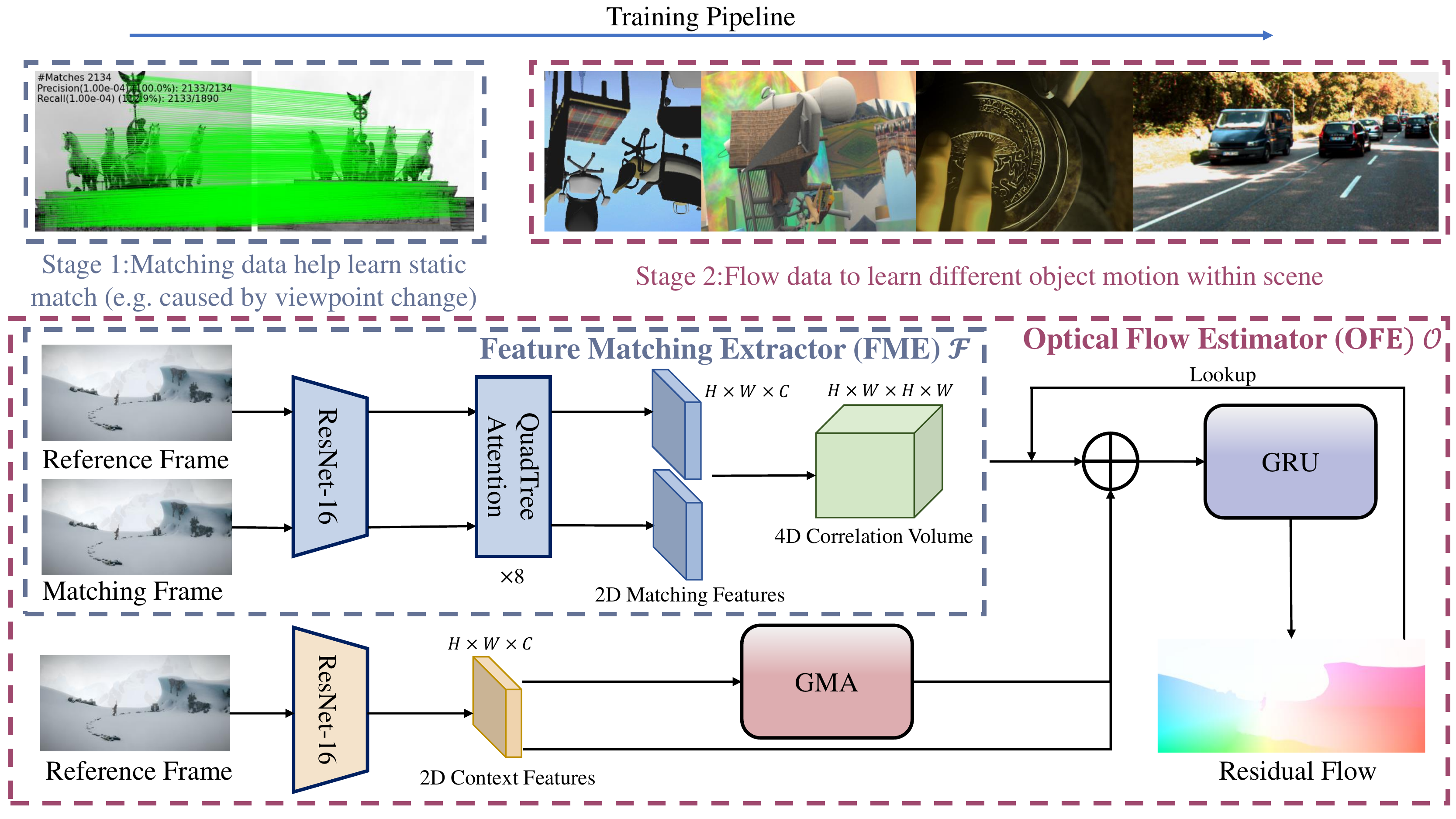}
\vspace{-0.1in}
\caption{Overview of our MatchFlow. The simplified training pipeline is shown at the top, while details of each stage are specifically listed below. $H, W$ indicate 1/8 height and width of the input image respectively. Here the GMA means global motion aggregation module~\cite{jiang2021learning}.}
\label{fig:overview}
\vspace{-0.15in}
\end{figure*}

\noindent\textbf{Pipeline Overview.} 
The training pipeline is shown in Fig.~\ref{fig:overview}, which includes two stages. For stage one, we first train the Feature Matching Extractor (FME) $\mathcal{F}$ on massive real-world GIM data to learn static scene matching caused by viewpoint changes (Sec.~\ref{sec:geometric_pretraining}). 
Thus given a pair of images $I_1, I_2$, the coarse 4D correlation volume can be achieved with $C=\mathcal{F}(I_1,I_2)$.
For stage two, we follow~\cite{ilg2017flownet, teed2020raft}, and jointly finetune the FME $\mathcal{F}$ with the specialized Optical Flow Estimator (OFE) $\mathcal{O}$ to get the final displacement field as $(f^1, f^2)=\mathcal{O}(\mathcal{F}(I_1,I_2))$ (Sec.~\ref{sec:of_refine}).
Stage two is gradually applied on FlyingChair, FlyingThings3D, Sintel, and KITTI with increasing difficulty. 
Such a progressive pipeline makes MatchFlow address intractable object motion within complex scenes.

% \noindent\textbf{Network.} We employ the successful RAFT~\cite{teed2020raft} and GMA~\cite{jiang2021learning} as our optical flow estimation network as shown in Fig~\ref{fig:overview}. Given $I_1$ and $I_2$, we firstly use feature extraction module extracting 2D matching features $F_1, F_2$ and calculate the inner product to get a multi-scale 4D correlation volume. And 8 QuadTree attention blocks~\cite{tang2022quadtree} are utilized in our feature extraction module to learn more discriminative feature within non-occluded regions. Furthermore, Global Motion Aggregation module (GMA)~\cite{jiang2021learning} are used to propagate accurate motion feature within non-occluded regions learned from matching data to occluded regions based on the self-similarity of 2D context features, which can successfully resolve ambiguities caused by occlusions. Finally, GRU can learn to lookup from the correlation volume and iteratively decode the motion feature and context feature into a series of residual flow. The sum of the residual flows is the final output flow.

\subsection{Geometric Image Matching Pre-training\label{sec:geometric_pretraining}}

First, we elaborate on the idea of training FME on GIM task in this section.
Previous methods~\cite{ilg2017flownet, teed2020raft, jiang2021learning, sui2022craft, zhao2022global} are trained on optical flow data directly, without learning the feature matching through static scenes. FlowNet2~\cite{ilg2017flownet} finds that not only the quality of data is important, but also the data presenting order during the optical flow training is critical, which is closely related to the curriculum learning~\cite{bengio2009curriculum}.
It is trained on FlyingChair first, then trained on FlyingThings3D, and finally fine-tuned on target data Sintel. Among them, FlyingChair is the simplest dataset, composed of 22k image pairs of chairs superimposed on randomly selected background images. Furthermore, FlyingChair only contains planar motions.
FlyingThings3D contains about 88k image pairs generated by rendering random scenes of 3D models from ShapeNet~\cite{savva2015semantically} in front of static 3D backgrounds, which manifests real 3D motion and lighting effects with more challenges.

However, as both FlyingChair and FlyingThings3D are synthetic datasets, there is a lack of authenticity. Moreover, motions among different objects or between foregrounds and backgrounds in FlyingChair are not consistent. Thus FlyingChair is still not the `simplest motion' for curriculum learning. Therefore, we propose a simpler scenario to better assist the early representation learning of the network: optical flow estimation of consistent matching scenes, which can be regarded as GIM.
Besides, GIM enjoys much larger real-world datasets with various displacement and appearance changes, \emph{e.g.}, MegaDepth~\cite{li2018megadepth} and Scannet~\cite{dai2017scannet} with reliable enough matching labels.
% as it's simper to label true matching point for image pairs given camera parameters and depth measures.

% , providing no labels for occluded region.
On the other hand, limited by the hardware~\cite{dai2017scannet}, algorithm precision, and data quality~\cite{li2018megadepth}, GIM data is usually labeled sparsely with only non-occluded regions.
So we only train FME on GIM data. Specifically, we extract the feature $F_1$ and $F_2$ from the reference frame and matching frame respectively on $1/8$ of input resolution. And stacked QuadTree attention blocks~\cite{tang2022quadtree} are incorporated into FME with self and cross-attention to enhance the feature learning between the image pair efficiently.
Such an interleaving self and cross-attention manner is critical for the final optical flow performance as verified in Tab.~\ref{tab:ablation}.
We then construct the 4D correlation volume $C$ as follows:
\begin{equation}
 C(i, j) = \langle F_1(i), F_2(j)\rangle\in\mathbb{R}^{H\times W\times H\times W},
\end{equation}
where $i, j$ indicate the index of feature map $F_1$ and $F_2$; $H, W$ indicate 1/8 height and width of the input image.

\noindent\textbf{Pre-training Loss.}
We apply the dual-softmax~\cite{rocco2018neighbourhood,tyszkiewicz2020disk} to get the probability before the optimization.
Specifically, the matching probability $\mathcal{P}_c$ is calculated as:
\begin{align}
\negthickspace\negthickspace\negthickspace
\mathcal{P}_c(i, j) = \mathrm{softmax} (C (i, \cdot) / \tau)_{j}\cdot\mathrm{softmax}(C ( \cdot, j) / \tau)_{i},
\end{align}
where temperature $\tau=0.1$. 
We get ground-truth corresponding matches $\mathcal{M}_c^{gt}$ through depth and camera poses from GIM data in 1/8-resolution.
% as the mutual visible corresponding points in 1/8-resolution grids. 
Our loss function $\mathcal{L}_c$ is:
\begin{align}
\mathcal{L}_c=-\frac{1}{|\mathcal{M}_c^{gt}|}\sum_{(i, j)\in\mathcal{M}_c^{gt}}\log \mathcal{P}_c(i, j).
\end{align}
We also use additional $l_2$ loss in 1/2-resolution for fine-grained supervision~\cite{sun2021loftr}.

\subsection{Optical Flow Refinement}
\label{sec:of_refine}

After the pre-training of FME on matching dataset, we then finetune our FME with OFE on FlyingChair, FlyingThings3D, Sintel, and KITTI following previous data schedule~\cite{ilg2017flownet, teed2020raft, jiang2021learning, sui2022craft, zhao2022global}.
We employ the successful RAFT~\cite{teed2020raft} and GMA~\cite{jiang2021learning} as our OFE as shown in Fig~\ref{fig:overview}. Given $I_1$ and $I_2$, we first use FME to extract 2D matching features $F_1, F_2$ and calculate the inner product to get a multi-scale 4D correlation volume. QuadTree attention blocks are utilized in FME to learn more discriminative features within non-occluded regions. Furthermore, the global motion aggregation module~\cite{jiang2021learning} is used to propagate accurate motion features within non-occluded regions learned from matching data to occluded 
ones based on the self-similarity of 2D context features, which can successfully resolve ambiguities caused by occlusions. Finally, GRU can learn to lookup from the correlation volume and iteratively decode the motion feature and context feature into a series of residual flows. Sum of the residual flows is the final output flow.
% we employ the RAFT~\cite{teed2020raft} and GMA~\cite{jiang2021learning} as our afterwards specialized optical flow modules. RAFT is convolutional GRU based network, which predicts a series of residual flow given the 4D correlation volume and context feature. We initialize the flow as zeros everywhere and the final output are the sum of these residual flows. Besides, GMA is a module designed specialized for propagating motion from non-occluded to occluded regions as there is ambiguities to match occluded regions on feature map. Benefited from our geometric image matching pre-training, we can get much higher correlation within non-occluded regions, and more accurate motion within non-occluded region can thus benefit occluded regions a lot.

\noindent\textbf{Optical Flow Loss.} Given the partially summed residual flows $\{f_1, \dots, f_N\}$, we supervise our network use $l_1$ loss following previous work~\cite{teed2020raft}:
\begin{equation}
\mathcal{L}=\sum_{i=1}^N\gamma^{N-i}||f_{gt}-f_i||_1,
\end{equation}
where $f_{gt}$ is the ground truth flow, $\gamma=0.8$ on FlyingChair and FlyingThings3D, and 0.85 on Sintel and KITTI.

\section{Experiments}

\noindent\textbf{Implementation Details.}
We utilize MegaDepth~\cite{li2018megadepth} as our matching dataset and train the feature encoder with randomly sampled 36,800 image pairs per epoch for a total of 30 epochs~\cite{sun2021loftr}. We then follow the standard training pipeline~\cite{teed2020raft, jiang2021learning} of first training our model on FlyingChair (C) for 120k iterations (batch size 8) and FlyingThings3D (T) for another 120k iterations (batch size 6). Following training on C+T, we finetune the model using data from clean and final passes of Sintel (S), KITTI (K), and HD1K (H).
% Our Feature encoder is a ResNet. And there is 8 QuadTree attention blocks within our feature extraction modules. The other modules follow the GMA~\cite{jiang2021learning} and RAFT~\cite{teed2020raft}. 
Our method is implemented with Pytorch~\cite{paszke2019pytorch} on 2 RTX 3090 GPUs. And we use the one-cycle learning rate~\cite{smith2019super}, setting the highest learning rate for FlyingChairs to $2.5\times 10^{-4}$ and the remaining to $1.25\times 10^{-4}$. We denote the full model as \textbf{MatchFlow(G)}, and the model without GMA module as \textbf{MatchFlow(R)}. 

% \subsection{Comparisons with Prior Works}

\subsection{Sintel} 
\begin{figure*}
\centering
\includegraphics[width=0.95\linewidth]{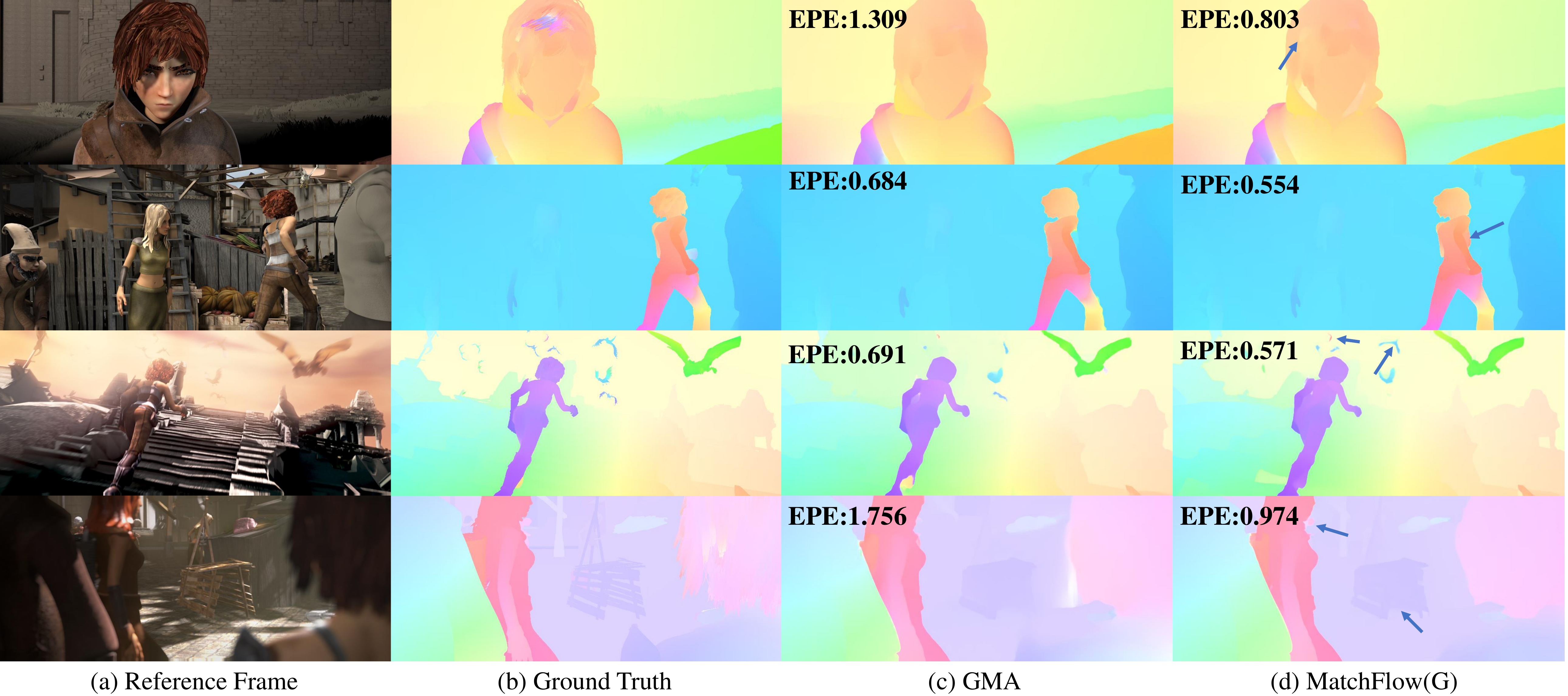}
\vspace{-0.1in}
\caption{Qualitative comparison on Sintel test set. First two rows are from clean pass, and the last two from final pass. Notable areas are pointed out by arrows. Please zoom in for details.\label{fig:sintel_qualitative}}
\end{figure*}

Following~\cite{teed2020raft}, we first evaluate the generalization error on the Sintel training set after training on C+T. As shown in Tab.~\ref{tab:main_results}, MatchFlow(G) achieves a 20.8\% reduction in Average End-Point-Error (AEPE) of Sintel clean pass from 1.30 to 1.03 based on GMA.
For AEPE of the final pass, MatchFlow(G) achieves a 10.6\% reduction from 2.74 to 2.45 based on GMA.
MatchFlow(R) also improves the RAFT by 20.3\% and 3.7\%. These results indicate the great generalization of our method. For the Sintel test set, MatchFlow(G) and MatchFlow(R) improve the GMA and RAFT by 11.5\% and 17.4\% on the clean pass. And MatchFlow(G) ranks first on Sintel clean pass among all published works. On Sintel final pass, MatchFlow(G) obtains an AEPE of 2.37 and outperforms GMA by 0.1 pixels. 
We also provide qualitative results on Sintel from the official website in Fig.~\ref{fig:sintel_qualitative}. The flow estimation from MatchFlow(G) exhibits clearer details (first, second, and fourth rows) and captures small and fast-moving birds in the third row.

\subsection{KITTI} 
For the KITTI training set, MatchFlow(G) reduces the generalization error EPE and Fl-all of GMA by 13.0\% and 8.8\% as shown in Tab.~\ref{tab:main_results}. MatchFlow(R) even achieves the best generalization performance with Fl-all 13.6\% with slightly worse EPE compared to MatchFlow(G). Fl-all here refers to the percentage of optical flow vectors whose EPE is larger than 3 pixels or 5\% of the ground truth magnitude.
Note that the Fl-all gap between MatchFlow(G) and MatchFlow(R) is somehow reasonable. As KPA-Flow~\cite{luo2022kpa} found that GMA's attention map contains many globally misleading clues. These globally misleading clues can lead to more large errors.
For the KITTI test set, MatchFlow(R) and MatchFlow(G) obtain Fl-all of 4.72\% and 4.63\% respectively, which reduce the error of the base model by 7.5\% and 10.1\%. Fig.~\ref{fig:kitti_qualitative} provides two qualitative examples from the KITTI test set. 
Compared with GMA, our MatchFlow(G) can successfully recognize the small signage in the first image and reduces the flow leakage around car boundaries in the second image. These results show the effectiveness and superiority of our GIM pre-training.

\begin{table*} \small
\centering
\vspace{-0.1in}
\caption{Quantitative comparison on standard benchmark. `A' indicates the AutoFlow~\cite{sun2021autoflow} dataset. `C+T': Succeeding training on FlyingChairs (C) and FlyingThings3D (T), we evaluate the capacity of generalization on Sintel (S) and KITTI (K) training sets. `C+T+S+K+H': Training samples from T, S, K, and HD1K (H) are included in our training set for further finetuning. Results on training set are shown in parentheses. The top and second-place results are bolded and underlined, respectively. $^\dagger$ indicates tile technique~\cite{jaegle2021perceiver}. And $^\star$ indicates evaluating with RAFT's ``warm-start" strategy.\label{tab:main_results}}
\vspace{-0.1in}
\begin{tabular}{clccccccc}
\toprule
\multirow{2}{*}{Training} & \multirow{2}{*}{Method} & \multicolumn{2}{c}{\uline{Sintel (train)}} & \multicolumn{2}{c}{\uline{KITTI-15 (train)}} & \multicolumn{2}{c}{\uline{Sintel (test)}} & \uline{KITTI-15 (test)}\tabularnewline
 &  & Clean & Final & Fl-epe & Fl-all & Clean & Final & Fl-all\tabularnewline
\hline
\multirow{2}{*}{A} & Perceiver IO~\cite{jaegle2021perceiver} & 1.81$^{\dagger}$ & 2.42$^{\dagger}$ & 4.98$^{\dagger}$ & - & - & - & -\tabularnewline
 & RAFT-A~\cite{sun2021autoflow} & 1.95 & 2.57 & 4.23 & - & - & - & -\tabularnewline
\hline
\multirow{14}{*}{C+T} & PWC-Net~\cite{sun2018pwc} & 2.55 & 3.93 & 10.35 & 33.7 & - & - & -\tabularnewline
 & FlowNet2~\cite{ilg2017flownet} & 2.02 & 3.54 & 10.08 & 30.0 & 3.96 & 6.02 & -\tabularnewline
 & RAFT~\cite{teed2020raft} & 1.43 & 2.71 & 5.04 & 17.4 & - & - & -\tabularnewline
 & Separable Flow~\cite{zhang2021separable} & 1.30 & 2.59 & 4.60 & 15.9 & - & - & -\tabularnewline
 & GMA~\cite{jiang2021learning} & 1.30 & 2.74 & 4.69 & 17.1 & - & - & -\tabularnewline
 & AGFlow~\cite{luo2022learning} & 1.31 & 2.69 & 4.82 & 17.0 & - & - & -\tabularnewline
 & KPA-Flow~\cite{luo2022kpa} & 1.28 & 2.68 & 4.46 & 15.9 & - & - & -\tabularnewline
 & DIP~\cite{zheng2022dip} & 1.30 & 2.82 & 4.29 & \uline{13.7} & - & - & -\tabularnewline
 & GMFlowNet~\cite{zhao2022global} & 1.14 & 2.71 & 4.24 & 15.4 & - & - & -\tabularnewline
 & GMFlow~\cite{xu2022gmflow} & 1.08 & 2.48 & 7.77 & 23.40 & - & - & -\tabularnewline
 & CRAFT~\cite{sui2022craft} & 1.27 & 2.79 & 4.88 & 17.5 & - & - & -\tabularnewline
 & FlowFormer~\cite{huang2022flowformer} & \textbf{1.01} & \textbf{2.40} & \uline{4.09}$^{\dagger}$ & {14.7}$^{\dagger}$ & - & - & -\tabularnewline
 & SKFlow~\cite{sun2022skflow} & 1.22 & 2.46 & 4.27 & 15.5 & - & - & -\tabularnewline
 & \textbf{MatchFlow(R) (Ours)} & 1.14 & 2.61 & 4.19$^{\dagger}$ & \textbf{13.6}$^{\dagger}$ & - & - & -\tabularnewline
 & \textbf{MatchFlow(G) (Ours)} & \uline{1.03} & \uline{2.45} & \textbf{4.08}$^{\dagger}$ & 15.6$^{\dagger}$ & - & - & -\tabularnewline
\hline
\multirow{11}{*}{C+T+S+K+H} 
 & PWC-Net+~\cite{sun2019models} & (1.71) & (2.34) & (1.50) & (5.3) & 3.45 & 4.60 & 7.72\tabularnewline
 & RAFT~\cite{teed2020raft} & (0.76) & (1.22) & (0.63) & (1.5) & 1.61$^{\star}$ & 2.86$^{\star}$ & 5.10\tabularnewline
 & RAFT-A~\cite{sun2021autoflow} & - & - & - & - & 2.01 & 3.14 & 4.78\tabularnewline
 & Separable Flow~\cite{zhang2021separable} & (0.69) & (1.10) & (0.69) & (1.60) & 1.50 & 2.67 & {4.64}\tabularnewline
 & GMA~\cite{jiang2021learning} & (0.62) & (1.06) & (0.57) & (1.2) & 1.39$^{\star}$ & 2.47$^{\star}$ & 5.15\tabularnewline
 & AGFlow~\cite{luo2022learning} & (0.65) & (1.07) & (0.58) & (1.2) & 1.43$^{\star}$ & 2.47$^{\star}$ & 4.89\tabularnewline
 & KPA-Flow~\cite{luo2022kpa} & (0.60) & (1.02) & (0.52) & (1.1) & 1.35$^{\star}$ & 2.36$^{\star}$ & \uline{4.60}\tabularnewline
 & DIP~\cite{zheng2022dip} & - & - & - & - & 1.44$^{\star}$ & 2.83$^{\star}$ & \textbf{4.21}\tabularnewline
 & GMFlowNet~\cite{zhao2022global} & (0.59) & (0.91) & (0.64) & (1.51) & 1.39 & 2.65 & 4.79\tabularnewline
 & GMFlow~\cite{xu2022gmflow} & - & - & - & - & 1.74 & 2.90 & 9.32\tabularnewline
 & CRAFT~\cite{sui2022craft} & (0.60) & (1.06) & (0.58) & (1.34) & 1.45$^{\star}$ & 2.42$^{\star}$ & 4.79\tabularnewline
 & FlowFormer~\cite{huang2022flowformer} & (0.48) & (0.74) & (0.53) & (1.11) & \uline{1.20} & \textbf{2.12} & 4.68$^{\dagger}$\tabularnewline
 & SKFlow~\cite{sun2022skflow} & (0.52) & (0.78) & (0.51) & (0.94) & 1.28$^{\star}$ & \uline{2.23}$^{\star}$ & 4.84\tabularnewline
 & \textbf{MatchFlow(R) (Ours)} & (0.51) & (0.81) & (0.59) & (1.3) & 1.33$^{\star}$ & 2.64$^{\star}$ & 4.72$^{\dagger}$ \tabularnewline
 & \textbf{MatchFlow(G) (Ours)} & (0.49) & (0.78) & (0.55) & (1.1) & \textbf{1.16}$^{\star}$ & {2.37}$^{\star}$ & {4.63}$^{\dagger}$\tabularnewline
\bottomrule
\end{tabular}
\vspace{-0.2in}
\end{table*}

\subsection{Ablation Studies}

We conduct a series of ablation studies on MatchFlow with training on C+T and evaluating on the test set of FlyingThings3D and the training set of Sintel and KITTI. The results are shown in Tab.~\ref{tab:ablation}. 
Since KITTI has a special aspect ratio, we perform more KITTI ablations in Tab.~\ref{tab:abla-kitti}.

\noindent\textbf{Geometric Image Matching Pre-training.} 
We first assess the effect of GIM pre-training with the same network structure (with 8 QuadTree attention blocks). As seen in Tab.~\ref{tab:ablation}, solely adding more parameters for FME only brings marginal improvement, and performs much worse than GMA on the Sintel final pass. With GIM pre-training, we can get much better performance on all datasets, except for the slightly worse Fl-all of KITTI, which validates the benefit of pre-training.

\noindent\textbf{Number of Attention Blocks.} 
We attempt to reduce the attention blocks in Tab.~\ref{tab:ablation}. When the number of attention blocks was reduced, the performance decreased sharply on Sintel while slowly on KITTI. When there are no attention blocks, our FME is a pure CNN-based network. Although FME is pre-trained with GIM, it performs even worse on Sintel than random initialized GMA. This may be caused by the network `overfitting' on the matching task. A recent work~\cite{park2022vision} suggests that attention layers can smooth out irregularities and high-frequency noises with a better generalization. We thus use 8 QuadTree attention blocks by default. Furthermore, to compare GMA and our model with aligned settings, we also provide the results of GMA-L in Tab.~\ref{tab:ablation}. GMA-L is a large version of GMA by doubling the channels. And the results of GMA-L are taken from~\cite{huang2022flowformer}. However, its performance is much worse than our model with more parameters.
To sum up, the new design of GIM pre-training and interleaved attention-based FME outperforms the one with trivially extended model scales.
% , \emph{e.g.}, a 29\% larger error than our model. 
% , is what has improved our model's performance, to sum up.

\noindent\textbf{Attention Type.} 
We also try Linear attention~\cite{katharopoulos2020transformers} and Global-Local Attention (GLA)~\cite{chen2022aspanformer} with comparable parameters in this subsection. Except for KITTI's Fl-all of GLA, Linear attention and GLA are inferior to QuadTree attention with much worse results on FlyingThings3D and cross-dataset generalization on Sintel.

\noindent\textbf{Global Matching.} 
Finally, we compare the technique proposed by GMFlowNet~\cite{zhao2022global}, where match initialization (Match Init.) refers to the usage of robust matching point based on 4D correlation volume to initialize the flow refinement during training and testing. Moreover, Match Loss indicates further optimizing the flow network with matching loss~\cite{rocco2018neighbourhood,tyszkiewicz2020disk} when trained on optical flow datasets. Tab.~\ref{tab:ablation} shows that after training on the GIM datasets, either adding match initialization or match loss on the flow dataset can not provide better performance.

\noindent\textbf{More Ablations on KITTI.} 
As KITTI owns a much smaller aspect ratio, it is difficult for our attention block to generalize to such images as verified in the first row of Tab.~\ref{tab:abla-kitti}.
So we use the tile technique proposed by Perceiver IO~\cite{jaegle2021perceiver}. Formally, we split the KITTI images into 2 smaller overlapping sub-images and weighted average the results on overlapping regions. The tile technique can reduce the Fl-all from 18.2\% to 16.8\%. Furthermore, recent multi-scale training methods~\cite{dong2022incremental, cao2022mvsformer, Chai2022Any} can be generalized to much higher resolution after being trained on multi-scale images with few large-scale ones. We thus try to train our model on several different resolution images to generalize better on KITTI. However, it only achieves marginal improvement with much more computational resources as shown in Tab.~\ref{tab:abla-kitti}. And we do not use it in our final model. 
Finally, we find that simply zero-padding test images discourages the final performance of attention-based models in KITTI.
Because we employ the QuadTree attention, the image resolution should be multiplied by 32 instead of 8 compared with previous methods. The traditional zero-padding brings much thicker black borders and degrades the performance a lot in extreme cases~\cite{sui2022craft}. So we use bilinear interpolation to adjust the resolution with the tile technique for better performance as shown in Tab.~\ref{tab:abla-kitti}.

\begin{table*} \small
\centering
% \vspace{-0.1in}
\caption{Ablation studies. Settings used in our final model are underlined. $^\star$ indicates a large version of GMA, results from~\cite{huang2022flowformer}.\label{tab:ablation}}
\vspace{-0.1in}
\begin{tabular}{ccccccccc}
\toprule
\multirow{2}{*}{Experiment} & \multirow{2}{*}{Method} & \multicolumn{2}{c}{\uline{Things (test)}} & \multicolumn{2}{c}{\uline{Sintel (train)}} & \multicolumn{2}{c}{\uline{KITTI-15 (train)}} & \multirow{2}{*}{Parameters}\tabularnewline
 &  & Clean & Final & Clean & Final & Fl-epe & Fl-all & \tabularnewline
\hline
\multirow{2}{*}{Baseline} & GMA~\cite{jiang2021learning} & 3.14 & 2.80 & 1.30 & 2.74 & 4.69 & 17.1 & 5.9M\tabularnewline
 & GMA-L$^{\star}$~\cite{huang2022flowformer} & - & - & 1.33 & 2.56 & 4.40 & 15.9 & 17.0M\tabularnewline
\hline
\multirow{2}{*}{Local Matching Pre-training} & No & 2.82 & 2.56 & 1.27 & 2.84 & 4.12 & \textbf{14.4} & 15.4M\tabularnewline
 & \uline{Yes} & \textbf{2.12} & \textbf{2.07} & \textbf{1.03} & \textbf{2.45} & \textbf{4.08} & 15.6 & 15.4M\tabularnewline
\hline
\multirow{3}{*}{Number of Attention Blocks} & 0 & 2.97 & 2.81 & 1.36 & 2.98 & 4.58 & 16.3 & 5.9M\tabularnewline
 & 4 & 2.65 & 2.40 & 1.18 & 2.79 & 4.55 & \textbf{15.6} & 9.1M\tabularnewline
 & \uline{8} & \textbf{2.12} & \textbf{2.07} & \textbf{1.03} & \textbf{2.45} & \textbf{4.08} & \textbf{15.6} & 15.4M\tabularnewline
\hline
\multirow{3}{*}{Attention Type} & Linear Atten. & 2.82 & 2.56 & 1.24 & 3.00 & 4.32 & 15.6 & 14.3M\tabularnewline
 & \uline{QuadTree Atten.} & \textbf{2.12} & \textbf{2.07} & \textbf{1.03} & \textbf{2.45} & \textbf{4.08} & 15.6 & 15.4M\tabularnewline
 & GLA & 2.66 & 2.55 & 1.24 & 2.82 & 4.29 & \textbf{15.0} & 18.7M\tabularnewline
\hline
\multirow{3}{*}{Global Matching} & Match Init. & 2.23 & 2.29 & 1.44 & 2.93 & 5.30 & 17.4 & 15.4M\tabularnewline
 & Match Loss & 2.22 & 2.29 & \textbf{1.00} & 2.50 & 4.78 & 16.1 & 15.4M\tabularnewline
 & \uline{No} & \textbf{2.12} & \textbf{2.07} & 1.03 & \textbf{2.45} & \textbf{4.08} & \textbf{15.6} & 15.4M\tabularnewline
\bottomrule
\end{tabular}
\vspace{-0.15in}
\end{table*}

\begin{table} \small
\centering
% \vspace{-0.1in}
\caption{More Ablations on KITTI using MatchFlow(G) trained on C+T. We use Resize+Tile in our final model.\label{tab:abla-kitti}}
\vspace{-0.1in}
\begin{tabular}{cccccc}
\toprule
\multicolumn{2}{c}{\uline{Resolution}} & \uline{Training} & \uline{Test} & \multicolumn{2}{c}{\uline{KITTI-15 (train)}}\tabularnewline
Pad & Resize & Multiscale & Tile & Fl-epe & Fl-all\tabularnewline
\hline
\CheckmarkBold{} &  &  &  & 5.59 & 18.2\tabularnewline
\CheckmarkBold{} &  &  & \CheckmarkBold{} & 4.67 & 16.8\tabularnewline
\CheckmarkBold{} &  & \CheckmarkBold{} &  & 5.43 & 18.0\tabularnewline
\CheckmarkBold{} &  & \CheckmarkBold{} & \CheckmarkBold{} & 4.61 & 16.5\tabularnewline
 & \CheckmarkBold{} & \CheckmarkBold{} &  & 4.35 & 15.7\tabularnewline
 & \CheckmarkBold{} & \CheckmarkBold{} & \CheckmarkBold{} & \textbf{4.04} & \textbf{15.2}\tabularnewline
 & \CheckmarkBold{} &  &  & 4.54 & 16.3\tabularnewline
 & \CheckmarkBold{} &  & \CheckmarkBold{} & \uline{4.08} & \uline{15.6}\tabularnewline
\bottomrule
\end{tabular}
\vspace{-0.15in}
\end{table}

\subsection{Where are Gains Coming from?}

\begin{figure}
\centering
\subfloat{
\includegraphics[width=0.45\linewidth]{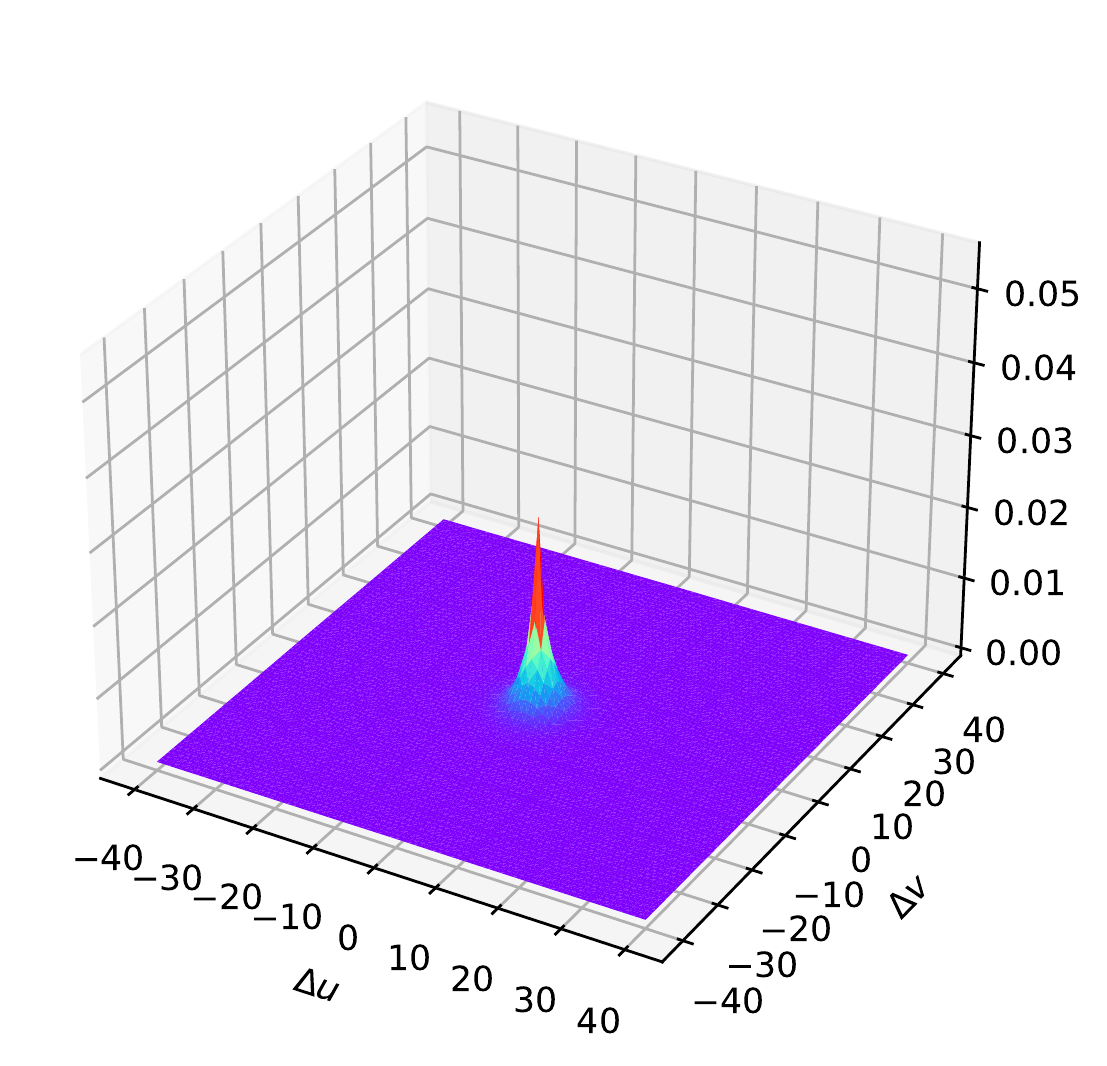}
}
\quad
\subfloat{
\includegraphics[width=0.45\linewidth]{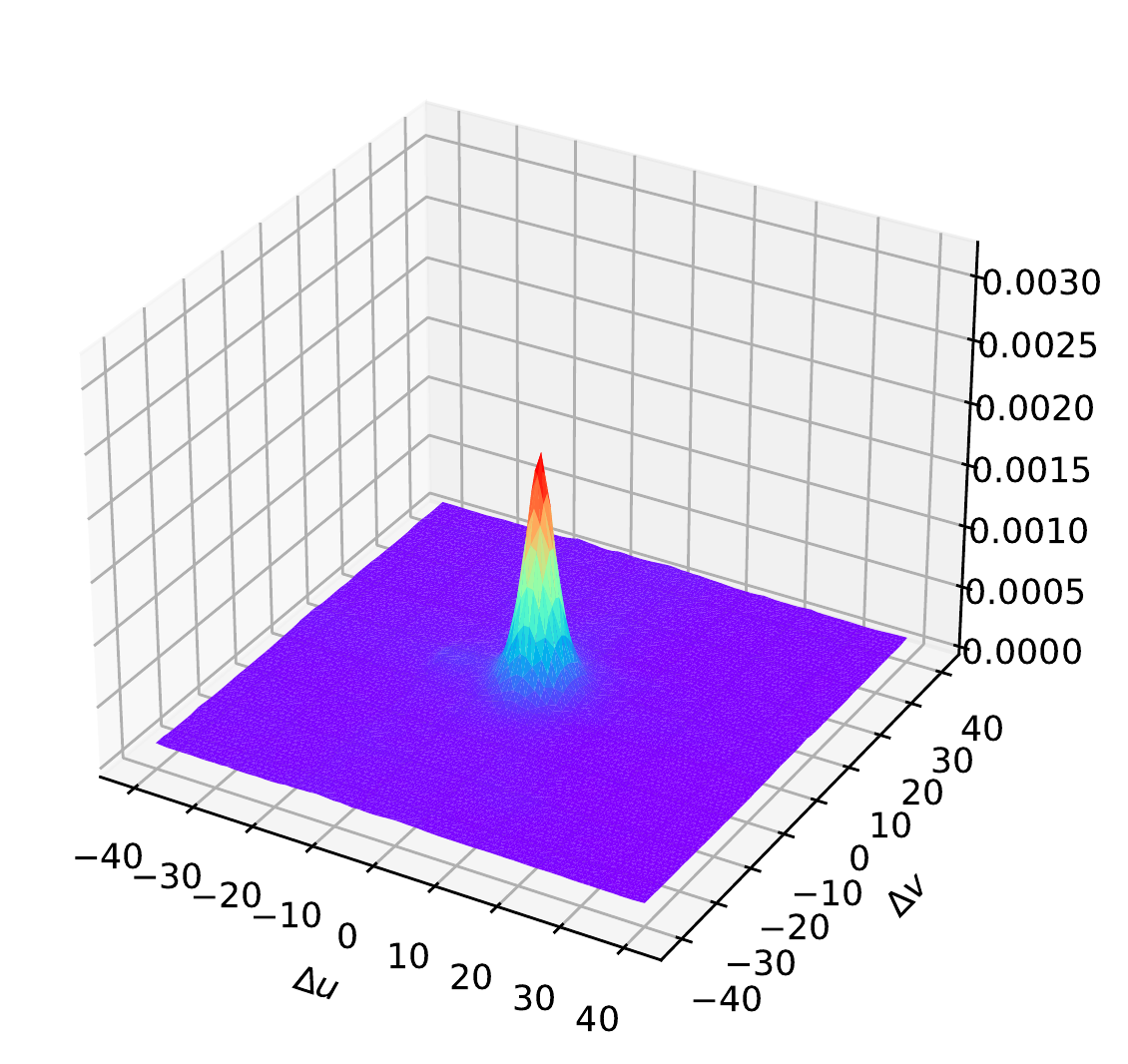}
}
\quad
\setcounter{subfigure}{0}
\subfloat[Noc]{
\includegraphics[width=0.45\linewidth]{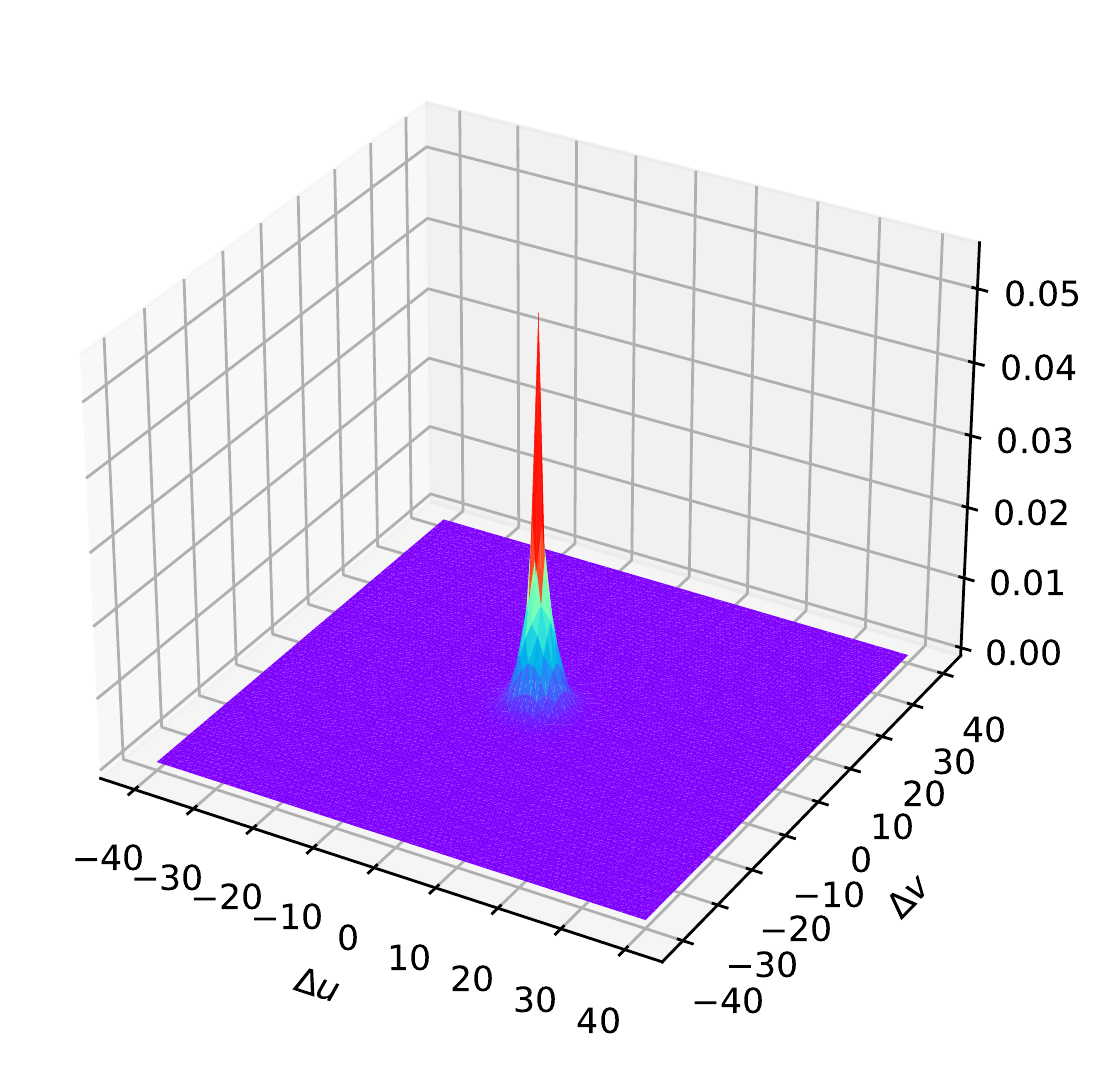}
}
\quad
\subfloat[Occ]{
\includegraphics[width=0.45\linewidth]{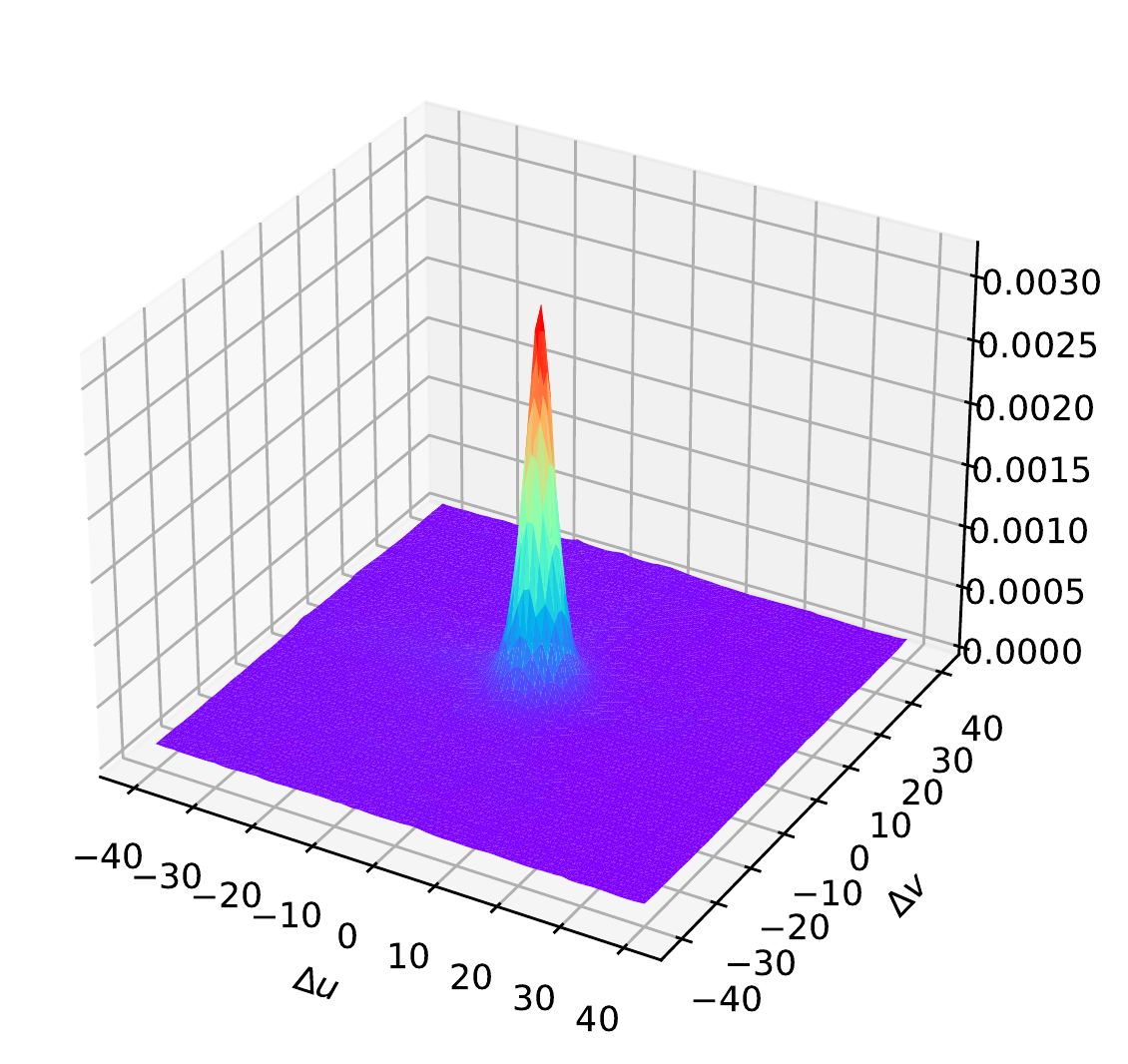}
}
% \vspace{-0.1in}
\caption{Visualizations of normalized correlation volumes within different region on 100 Sintel final pass images. The first row is for GMA, and the second row is MatchFlow(G). $(\Delta u, \Delta v)$ are the offsets away from ground truth flow. The center value (corresponding to ground truth flow) is $2.0\times$ and $1.6\times$ higher for our method vs. GMA within Noc and Occ regions respectively.\label{fig:cost_volume_visu}}
\vspace{-0.25in}
\end{figure}

\noindent\textbf{Occlusion Analysis.} To determine where our gains come from, we use the occlusion map provided by Sintel on the training set to divide the images into occluded (`Occ') and non-occluded (`Noc') areas. Following GMA~\cite{jiang2021learning}, we further divide the occluded area into in-frame (`Occ-in') and out-of-frame (`Occ-out') occlusions.
Using the model trained on FlyingThings3D, we evaluate it on three rendering passes of Sintel. As shown in Tab.~\ref{tab:sintel_analysis}, the relative improvement of our method compared to the baseline model, \emph{i.e.}, GMA, is mainly attributed to the error reduction within the Noc region. 
% Besides, benefiting from our high-quality 4D correlation volume, our MatchFlow(R) also achieves more improvement within the Occ region compared to MatchFlow(G). This is because that GMA has a specialized global motion aggregation module target the flow estimation within Occ regions. And the more difficult Occ-out occlusion even achieves the largest relative improvement in several passes. 
We further visualize the correlation volume on 100 Sintel final pass images with the model trained on FlyingThings3D in Fig.~\ref{fig:cost_volume_visu}. The first row is for GMA, and the second row is MatchFlow(G). The much higher central value of our method validates that GIM pre-training can benefit optical flow a lot, especially within the Noc region.

\begin{table} \small
\centering
% \vspace{-0.1in}
\caption{Optical flow error within occluded (`Occ') and non-occluded (`Noc') regions for three Sintel datasets. Each dataset's top outcomes and biggest relative improvement are shown in bold. Models are trained using the FlyingThings3D training set.\label{tab:sintel_analysis}}
\vspace{-0.1in}
\begin{tabular}{ccccc}
\toprule
Sintel &  & GMA & MatchFlow(G) & Rel. Impr.\tabularnewline
Pass & Type & (AEPE) & (AEPE) & (\%)\tabularnewline
\hline
\multirow{5}{*}{Clean} & Noc & 0.58 & \textbf{0.44} & \textbf{24.1}\tabularnewline
 & Occ & 10.58 & \textbf{8.51} & 19.6\tabularnewline
 & Occ-in & 7.68 & \textbf{6.42} & 16.4\tabularnewline
 & Occ-out & 12.52 & \textbf{9.63} & 23.1\tabularnewline
 & All & 1.30 & \textbf{1.03} & 20.8\tabularnewline
\hline
\multirow{5}{*}{Final} & Noc & 1.72 & \textbf{1.46} & 15.1\tabularnewline
 & Occ & 17.33 & \textbf{15.00} & 13.4\tabularnewline
 & Occ-in & 14.96 & \textbf{12.31} & \textbf{17.7}\tabularnewline
 & Occ-out & 16.44 & \textbf{15.32} & 6.8\tabularnewline
 & All & 2.74 & \textbf{2.45} & 10.6\tabularnewline
\hline
\multirow{5}{*}{Albedo} & Noc & 0.48 & \textbf{0.37} & \textbf{22.9}\tabularnewline
 & Occ & 9.54 & \textbf{7.98} & 16.4\tabularnewline
 & Occ-in & 7.51 & \textbf{6.37} & 15.2\tabularnewline
 & Occ-out & 10.22 & \textbf{8.39} & 17.9\tabularnewline
 & All & 1.15 & \textbf{0.92} & 20.0\tabularnewline
\bottomrule
\end{tabular}
\vspace{-0.15in}
\end{table}

\noindent\textbf{Real-world Dataset based Training.} We futher supply a comparison of generalization error to Depthstill~\cite{Aleotti2021Learning}, self-supervised pre-trained DINO feature~\cite{Caron2021DINO}, and class-supervised Twins-SVT feature~\cite{Chu2021twins}. We follow Depthstill~\cite{Aleotti2021Learning} to generate 20K virtual images and flow maps on MegaDepth (dubbed \textbf{dMegaDepth}) to finetune the model. Our method still performs best among them in more complex Sintel. And feature encoders trained by DINO~\cite{Caron2021DINO} and classification also perform worse than ours on both Sintel and KITTI. In contrast, the motion types of objects in KITTI are much simpler than Sintel, such as mostly forward translations or steerings. Depthstill~\cite{Aleotti2021Learning} greatly benefited from the last dense matching-based flow training on KITTI, while greatly degrading the results on Sintel.

\begin{table} \small
\centering
\caption{Generalization error of real-world dataset based training.\label{tab:dataset_training}}
\vspace{-0.1in}
\renewcommand\tabcolsep{4pt}
\begin{tabular}{cccccc}
\toprule
\multirow{2}{*}{Method} & \multirow{2}{*}{Dataset} & \multicolumn{2}{c}{\uline{Sintel}} & \multicolumn{2}{c}{\uline{KITTI-15}}
\tabularnewline
 &  & Clean & Final & Fl-epe & Fl-all\tabularnewline
\hline
{Baseline} & {C+T} & {1.27} & {2.84} & {4.12} & {14.4}\tabularnewline
{Depthstill} & {C+T+dMegaDepth} & {2.10} & {3.47} & \textbf{3.41} & \textbf{10.9}\tabularnewline
{DINO} & {ImageNet+C+T} & {1.52} & {3.08} & {5.50} & {19.2}\tabularnewline
{Twins-SVT} & {ImageNet+C+T} & {1.15} & {2.73} & {4.98} & {16.8}\tabularnewline
{Ours} & {MegaDepth+C+T} & \textbf{1.03} & \textbf{2.45} & {4.08} & {15.6}\tabularnewline
\bottomrule
\end{tabular}
\vspace{-0.1in}
\end{table}

\subsection{Parameters, Timing, and GPU Memory}

We show the computational overhead of our models in Tab.~\ref{tab:timing}. The parameter numbers of MatchFlow(R) and MatchFlow(G) are 14.8M and 15.4M respectively, which mainly attribute to 8 QuadTree attention blocks. We also test the GPU training memory on FlyingThings3D with 2 48G RTX A6000 GPUs. Since FlowFormer~\cite{huang2022flowformer} is trained on FlyingThings3D with image size $432\times 960$ to get better cross-dataset performance on Sintel, we can only support batch size 2 per GPU with peak memory consumption 43.7G. And the training image size of GMA and our models are $400\times 720$ and $416\times 736$ (as our resolution should be multiplied by 32 for QuadTree attention) respectively. Testing with batch size 3 per GPU, our models consume around 22.1 and 23.6G memory. And our training time is around 58 hours. We test the inference time on Sintel with a single RTX A6000 GPU. GMA takes 74ms on average while MatchFlow(G) takes 126ms per image pair. The GRU iteration number is set to 12 for all models. The overall computational overhead of MatchFlow is modest compared to FlowFormer, while our performance is still competitive.

\begin{table} \small
\centering
% \vspace{-0.1in}
\caption{Parameters, inference and training time, and memory.\label{tab:timing}}
\vspace{-0.1in}
\begin{tabular}{ccccc}
\toprule
Model & Para. & Infer. & Train. & Memory\tabularnewline
\hline
GMA~\cite{jiang2021learning} & 5.9M & 74 ms & 30 h & 11.7 G\tabularnewline
FlowFormer~\cite{huang2022flowformer} & 18.2 M & 149 ms & 107 h & 43.7 G\tabularnewline
MatchFlow(R) & 14.8 M & 110 ms & 53 h & 22.1 G\tabularnewline
MatchFlow(G) & 15.4M & 126 ms & 58 h & 23.6 G\tabularnewline
\bottomrule
\end{tabular}
\vspace{-0.15in}
\end{table}

\subsection{Failure Cases and Limitations\label{sec:limit}}
\begin{figure}
\centering
\includegraphics[width=0.98\linewidth]{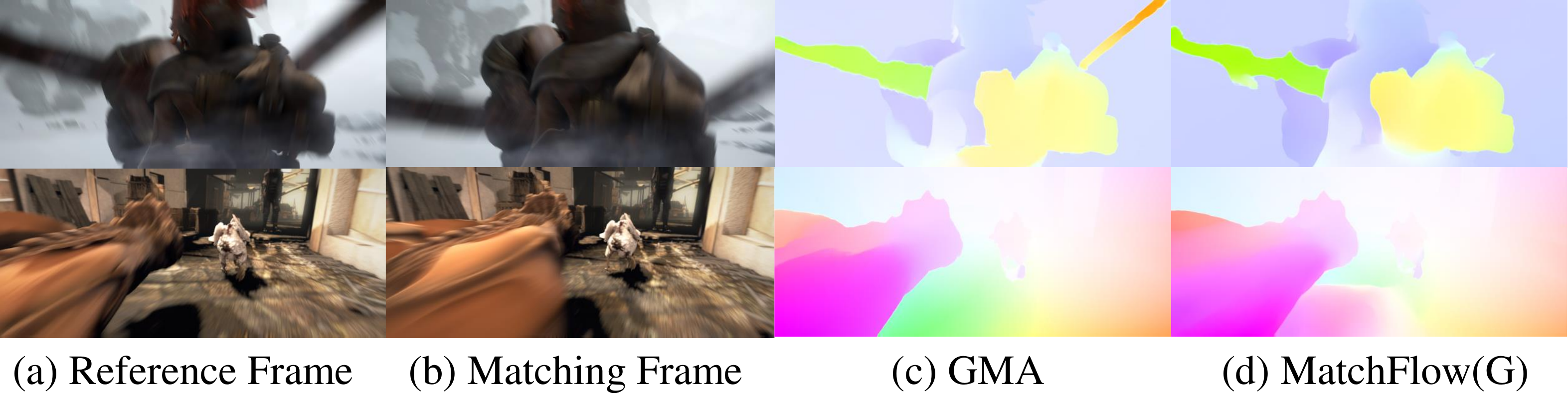}
\vspace{-0.05in}
\caption{Failure cases on Sintel final pass.\label{fig:failure_case}}
\vspace{-0.2in}
\end{figure}

%  and blurred areas are  matched
Though we achieve impressive results on KITTI and Sintel, we fail to generalize well to some extreme situations. Fig.~\ref{fig:failure_case} shows two failed cases. In the first case, our model ignores the right part of the weapon caused by motion blur. And the motion of shadow is estimated mistakenly in the second case. We conjecture that this is caused by inadequate training on such blur data, which discourages the performance of attention blocks. We believe that training on more diverse final-type data can improve the results.

\section{Conclusion}

This paper proposed to use geometric image matching as the pre-training task for learning more robust matching features for optical flow estimation. Sharing some common challenges with optical flow estimation while with much more real-world data, geometric image matching pre-training can help learn fundamental feature correlations of scenes. To improve the feature matching extractor, we further proposed adding stacked QuadTree attention layers for self and cross-attention learning. Extensive experiments showed great cross-dataset generalization and superior performance of our models. Future works may focus on reducing the attention layers in feature matching extractor while preserve competitive performance.

%%%%%%%%% REFERENCES
{\small
\bibliographystyle{ieee_fullname}
\bibliography{egbib}
}

\clearpage
\section*{Appendix}
\appendix

%%%%%%%%% BODY TEXT

\section{Details of QuadTree Attention}

In the Feature Matching Extractor (FME), we employ the QuadTree attention~\cite{tang2022quadtree} to enhance the feature. Specifically, given the image feature $F_1, F_2$ from ResNet-16, 8 stacked QuadTree attention blocks (4 self-attention and 4 cross-attention blocks) are incorporated into FME to enhance $F_1$ and $F_2$. We first linear project $F_1$ and $F_2$ to query $Q$, key $K$, and value $V$. Take cross-attention as an example:
\begin{align}
Q&=W_qF_1,\\
K&=W_kF_2,\\
V&=W_vF_2,
\end{align}
where $W_q, W_k, W_v$ are learnable parameters. We then construct 3-level pyramids for query $Q$, key $K$, and value $V$ by average pooling. After computing attention scores at the coarse level:
\begin{equation}
A = \mathrm{Softmax}(\frac{QK^T}{\sqrt{C}}),
\end{equation}
we select the top k key tokens with the highest attention scores for each query token. At the finer level, query sub-tokens only need to be evaluated with those key sub-tokens that correspond to one of the selected k key tokens at the coarse level. This process is repeated until reaches the finest level. We finally weighted average all selected value tokens at all levels through learnable weight and attention scores. And k is set to 16 for the coarsest level, and 8 for the remaining levels.

\section{Tile Technique}

As KITTI owns a much smaller aspect ratio, we use the tile technique~\cite{jaegle2021perceiver, huang2022flowformer}. Specifically, given a test image with size $(H_{test}, W_{test})$, we split it into several patches according to training image size $(H_{train}, W_{train})$. For example, it results in two patches starting at $(0, 0)$ and $(0, W_{test}-W_{train})$ if $H_{test}\leq H_{train}$; and four patches starting at $(0, 0)$, $(H_{test}-H_{train}, 0)$, $(H_{test}-H_{train}, W_{test}-W_{train})$, and $(0, W_{test}-W_{train})$ otherwise. For pixels covered by several patches, we weighted average the flows from these patches and get the final results. The weight is computed from the pixel's normalized distances $d_{u,v}$ to the corresponding patch center:
\begin{equation}
d_{u,v} = ||(u/H_{train}-0.5, v/W_{train}-0.5)||_2,
\end{equation}
where $(u, v)$ is the pixel's 2D index within each patch. And we use the Gaussian probability density function to get the final weight for each patch:
\begin{equation}
w_{u,v} = \frac{1}{\sqrt{2\pi}\sigma}\exp(-\frac{d_{u,v}^2}{2\sigma^2}),
\end{equation}
where $\sigma=0.05$.

\section{How much does the Megadepth pretraining provide a good starting point?}

\begin{figure}
\centering
\includegraphics[width=0.95\linewidth]{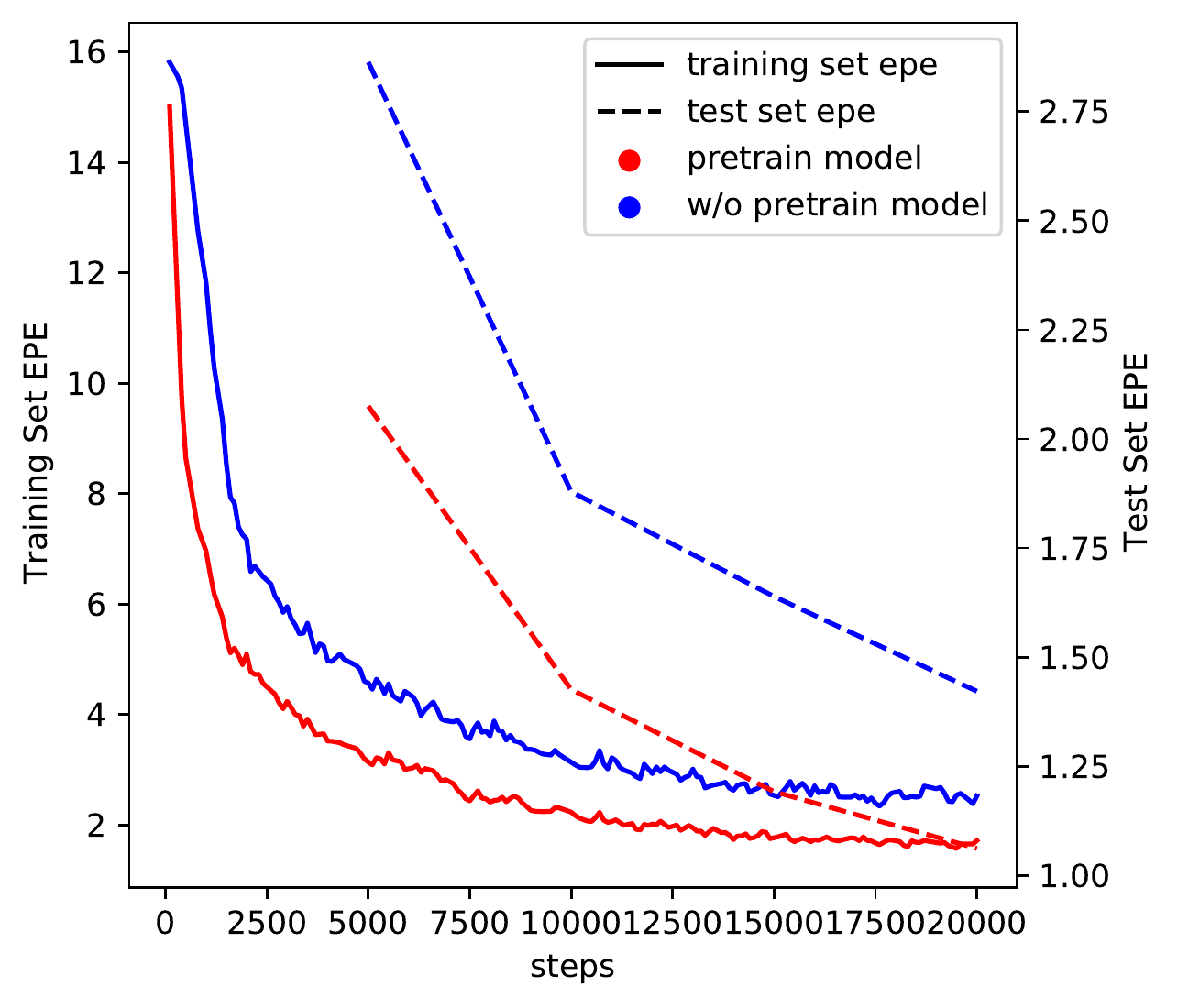}
\caption{EPE graph at early iterations on C+T pre-training.\label{fig:reburral_epe_early_iter_c+t}}
\end{figure}

We further provide an EPE graph at early iterations on C+T pre-training, for the Megadepth-trained model and a model from the scratch. Pretrain indeed provides a much better starting point and converges to lower error on training/test set as shown in Fig.~\ref{fig:reburral_epe_early_iter_c+t}.

Besides, as we only pretrain feature encoder by GIM, flow decoder is still learned from scratch on flow data. Directly finetuning model on C+T+S+K+H can result in poor performance (1.63 and 2.77 on Clean and Final of Sintel test set respectively) and serious grid artifacts around motion boundary. So the synthetic dataset pretraining (C+T) is still necessary for our method.

\section{More Qualitative Comparison}

\begin{figure*}
\centering
\includegraphics[width=0.95\linewidth]{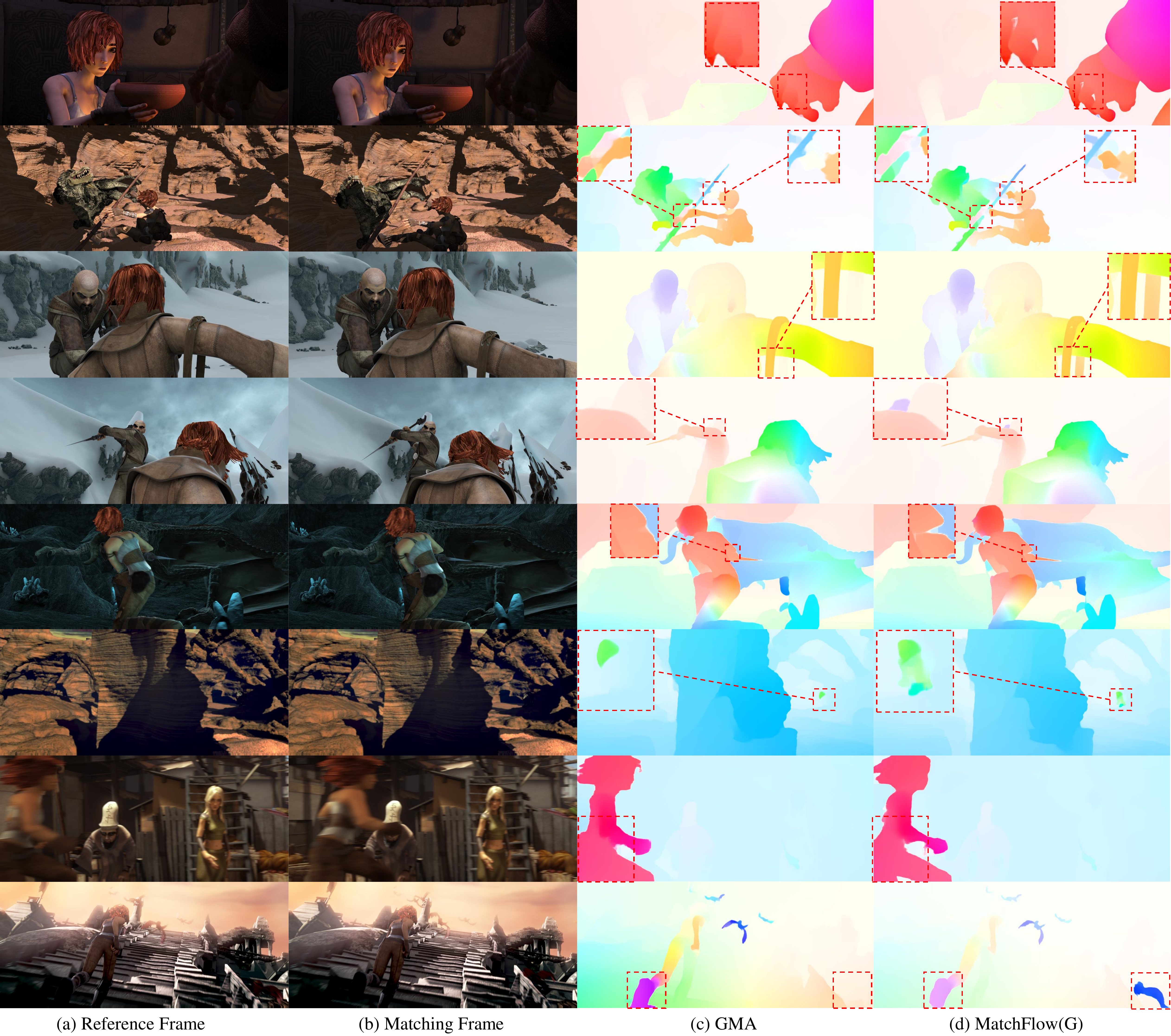}
\caption{More qualitative results on Sintel test set. First four rows are from clean pass, and the last four from final pass. Ground-truth optical flows are not available and are not shown. Red dashed boxes mark the regions of substantial improvements. Please zoom in for details.\label{fig:sintel_qualitative_supp}}
\end{figure*}

\begin{figure*}
\centering
\includegraphics[width=0.98\linewidth]{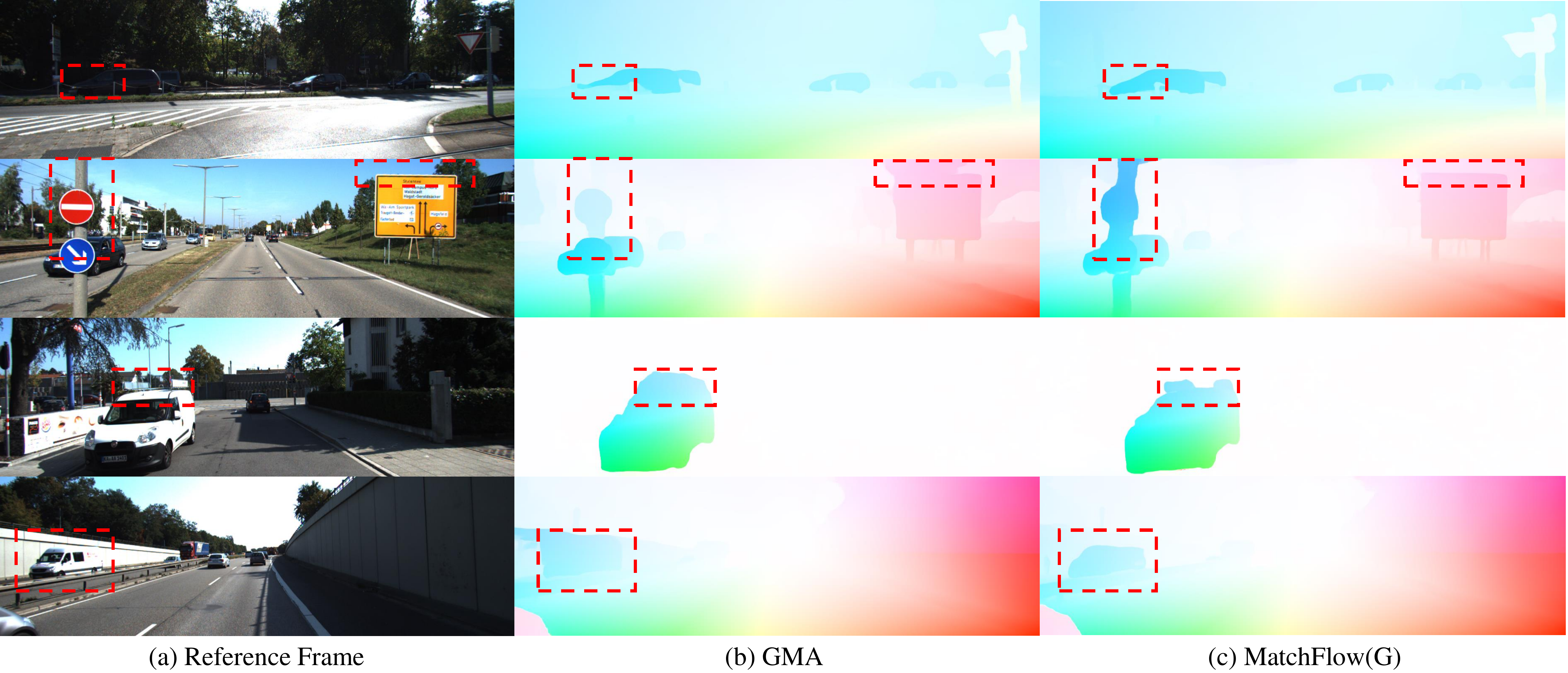}
\caption{More qualitative results on KITTI test set. Red dashed boxes mark the regions of substantial improvements. Please zoom in for details.\label{fig:kitti_qualitative_supp}}
\end{figure*}

More qualitative results on Sintel test set and KITTI set compared between our MatchFlow(G) and GMA~\cite{jiang2021learning} are given in Fig.~\ref{fig:sintel_qualitative_supp} and Fig.~\ref{fig:kitti_qualitative_supp}. As these samples from Sintel test set have no ground-truth optical flows, we can not give the AEPE and replace the ground-truth flows with matching frames in the second column in Fig.~\ref{fig:sintel_qualitative_supp}. We highlight the areas where our MatchFlow(G) beats GMA~\cite{jiang2021learning}. Please zoom in for more details.

\begin{figure*}
\centering
\includegraphics[width=0.98\linewidth]{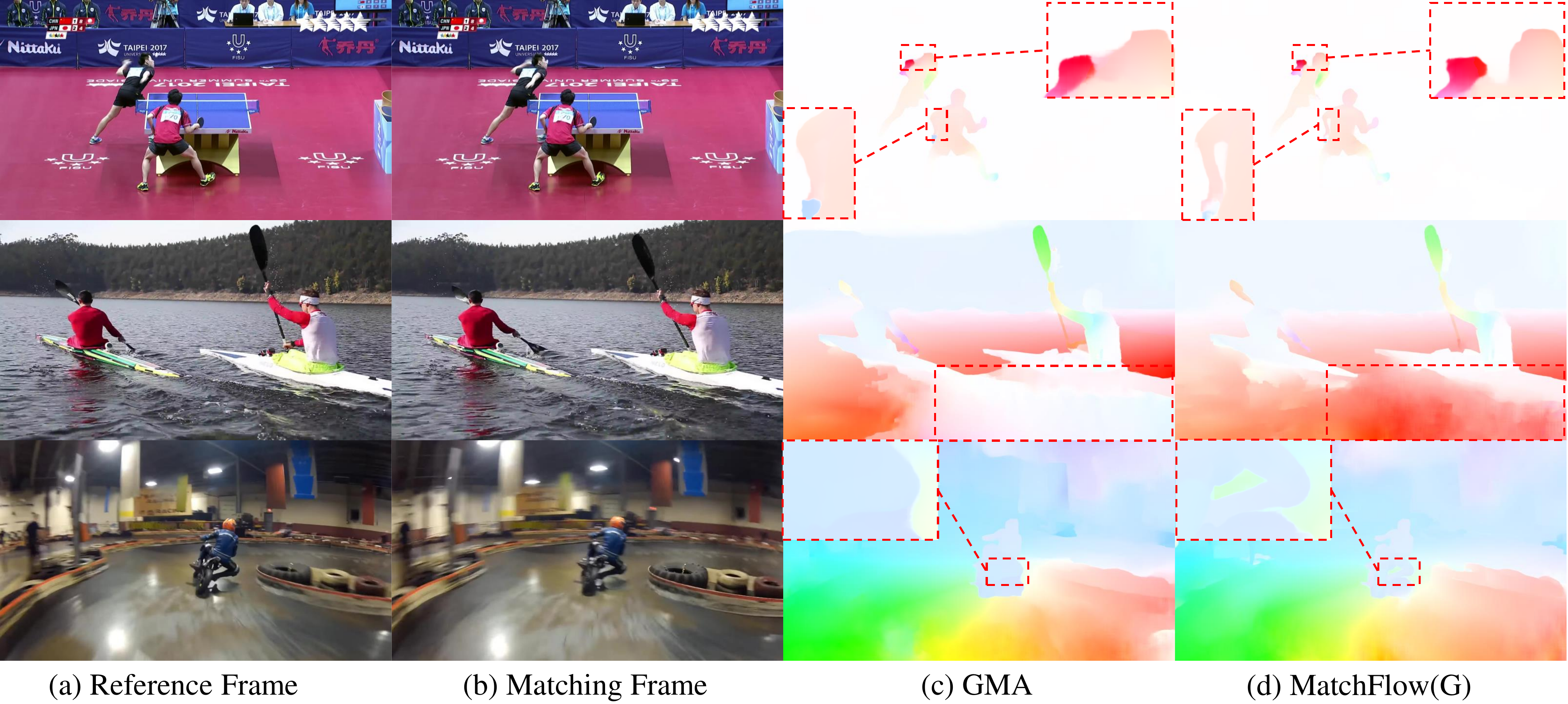}
\caption{Qualitative results on 1080p (1088x1920) DAVIS~\cite{Caelles_arXiv_2019} test set. Red dashed boxes mark the regions of substantial improvements. Please zoom in for details.\label{fig:davis_qualitative_supp}}
\end{figure*}

In addition, we provide qualitative comparison with GMA~\cite{jiang2021learning} on HD video from DAVIS~\cite{Caelles_arXiv_2019} test set. We test models on 1080p (1088x1920) resolution video and set the GRU iterations to 12 for both models. We do not use tile technique~\cite{jaegle2021perceiver} here. Both models are trained on Sintel. Fig.~\ref{fig:davis_qualitative_supp} shows that our model exhibits clearer details (first and third rows) and performs better on textureless regions (second row). Please zoom in for more details.

\section{Method of Correlation Volume Visualization}

We visualize the correlation volume following GMFlowNet~\cite{zhao2022global}. Specifically, given 4D correlation volume: $C\in\mathbb{R}^{H\times W\times H\times W}$, where $i, j$ indicate the index of feature map $F_1$ and $F_2$; $H, W$ indicate 1/8 height and width of the input image, we extract the local correlation map $F_i$ for point $i=(u,v)$ around the ground-truth optical flow $f_{gt}=(f_{gt}^1, f_{gt}^2)$ as follows:
\begin{align}
F_i = C(i, (u+f_{gt}^1+x, &v+f_{gt}^2+y))\in\mathbb{R}^{1\times 1\times 11\times 11}, \notag\\
&-5\leq x\leq 5, -5\leq y\leq 5.
\end{align}
As $H, W$ indicate 1/8 height and width of the input image, a local $11\times 11$ window in $C$ corresponds to $88\times 88$ local window in input image. We then normalize the local correlation map by \textit{Softmax}:
\begin{equation}
\hat{F}_i = \mathrm{Softmax}(F_i).
\end{equation}
Finally, we average $\hat{F}_i$ on all points within different
region on 100 Sintel final pass images. The results are shown in Fig.~4 of main paper.

\section{Screenshots of Sintel and KITTI Results}

\begin{figure*}
\centering
\includegraphics[width=0.98\linewidth]{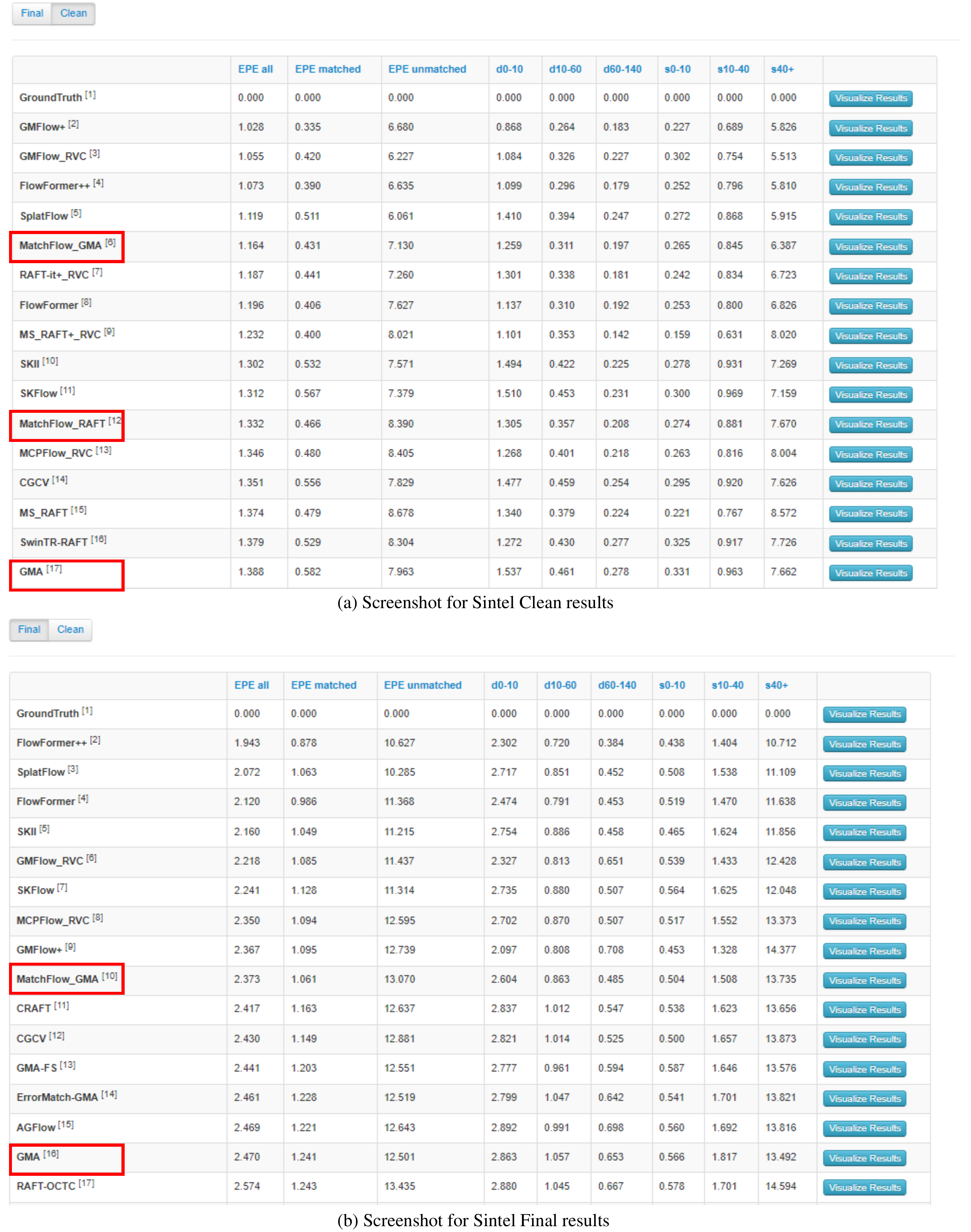}
\caption{Screenshots for Sintel Clean and Final results on the test server.\label{fig:screenshots_sintel}}
\end{figure*}

\begin{figure*}
\centering
\includegraphics[width=0.98\linewidth]{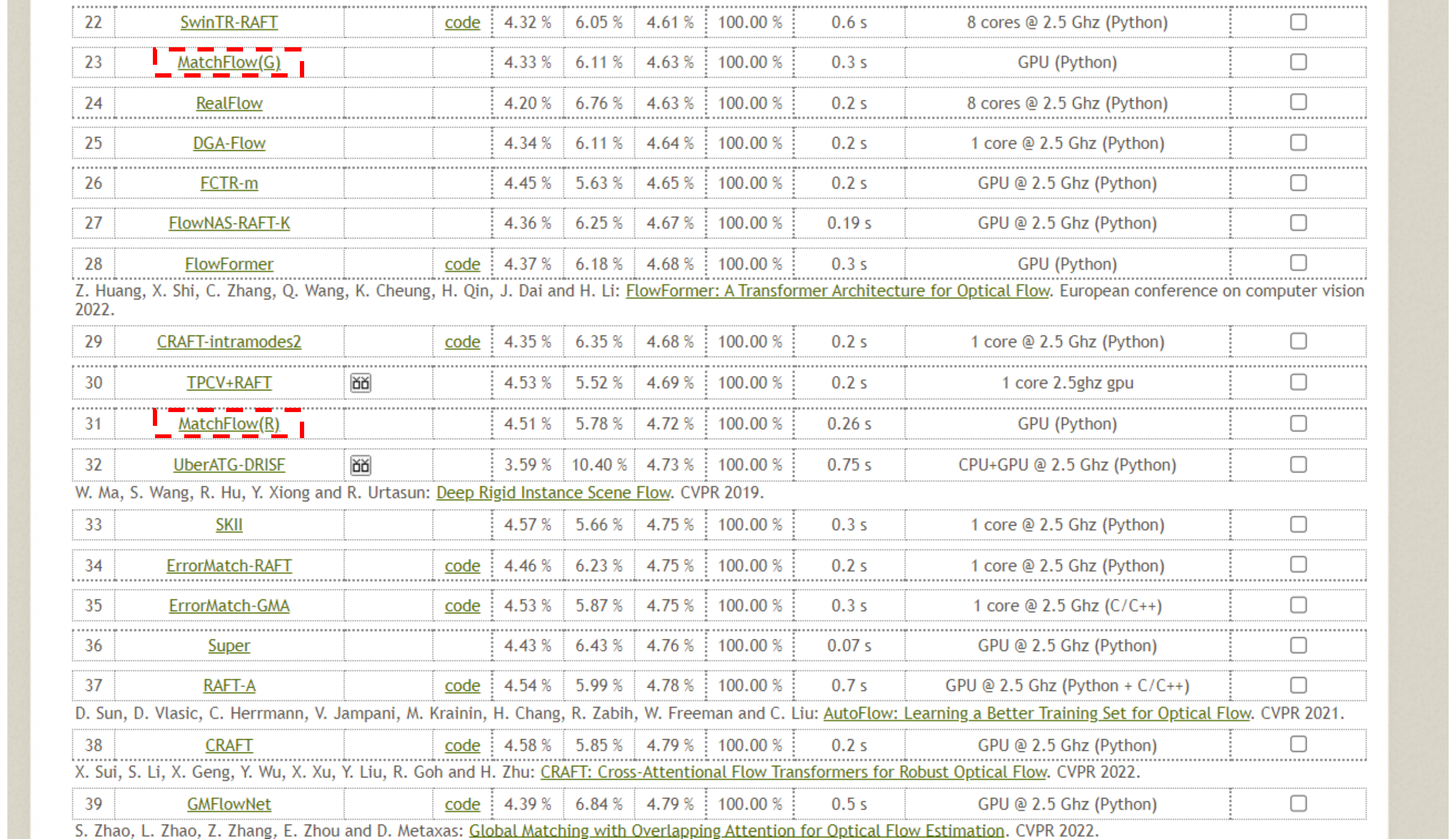}
\caption{Screenshots for KITTI optical flow evaluation 2015 results on the test server.\label{fig:screenshots_kitti}}
\end{figure*}

We provide anonymous screenshots of Sintel and KITTI results on the test sever in Fig.~\ref{fig:screenshots_sintel} and Fig.~\ref{fig:screenshots_kitti}. Our method ranks first on Sintel Clean pass and sceond on Sintel Final pass among all published approaches. Besides, we also achieve great performance improvement on KITTI test set. These results signifies the effectiveness of our approach.

\end{document}